\documentclass{article}

\usepackage{microtype}
\usepackage{graphicx}
\usepackage{subcaption}
\usepackage{booktabs} % for professional tables
\usepackage{multirow}

\usepackage{hyperref}

\ifdefined\icmlanonymous
    \usepackage{icml2026}
\else
    \usepackage[preprint]{icml2026}
\fi

\usepackage{amsmath}
\usepackage{amssymb}
\usepackage{mathtools}
\usepackage{amsthm}

\usepackage{algorithm}
\usepackage{algorithmic}
\usepackage[capitalize,noabbrev]{cleveref}

\theoremstyle{plain}
\newtheorem{theorem}{Theorem}[section]

\theoremstyle{definition}
\newtheorem{definition}[theorem]{Definition}

\theoremstyle{remark}

\usepackage[textsize=tiny]{todonotes}

\icmltitlerunning{Personalized ECG Generation Using Controllable Diffusion Model}

\def\modelname{ECGTwin}
\newcommand{\ms}[2]{#1$\pm\scriptstyle{#2}$}

\begin{document}

\twocolumn[
  \icmltitle{\modelname: Personalized ECG Generation Using Controllable Diffusion Model}



  \icmlsetsymbol{equal}{*}

  \begin{icmlauthorlist}
    \icmlauthor{Yongfan Lai}{1,2,3}
    \icmlauthor{Bo, Liu}{1,2}
    \icmlauthor{Xinyan Guan}{3}
    \icmlauthor{Qinghao Zhao}{4}
    \icmlauthor{Hongyan Li}{1,2}
    \icmlauthor{Shenda Hong}{3}
  \end{icmlauthorlist}

  \icmlaffiliation{1}{State Key Laboratory of General Artificial Intelligence, Beijing, China}
  \icmlaffiliation{2}{School of Intelligence Science and Technology, Peking University, Beijing, China}
  \icmlaffiliation{3}{National Institute of Health Data Science, Peking University, Beijing, China}
  \icmlaffiliation{4}{Department of Cardiology, Peking University People’s Hospital, Beijing, China}

  \icmlcorrespondingauthor{Shenda Hong}{hongshenda@pku.edu.cn}

  \icmlkeywords{Machine Learning, ICML}

  \vskip 0.3in
]



\printAffiliationsAndNotice{}  % no special notice (required even if empty)

\begin{abstract}
Personalized electrocardiogram (ECG) generation is to simulate a patient's ECG digital twins tailored to specific conditions.
It has the potential to transform traditional healthcare into a more accurate individualized paradigm, while preserving the key benefits of conventional population-level ECG synthesis.
However, this promising task presents two fundamental challenges: extracting individual features without ground truth and injecting various types of conditions without confusing generative model. 
In this paper, we present \textbf{\modelname}, a two-stage framework designed to address these challenges.
In the first stage, an \textit{Individual Base Extractor} trained via contrastive learning robustly captures personal features from a reference ECG. In the second stage, the extracted individual features, along with a target cardiac condition, are integrated into the diffusion-based generation process through our novel \textit{AdaX Condition Injector}, which injects these signals via two dedicated and specialized pathways.
Both qualitative and quantitative experiments have demonstrated that our model can not only generate ECG signals of high fidelity and diversity by offering a fine-grained generation controllability, but also preserving individual-specific features. Furthermore, \modelname\ shows the potential to enhance ECG auto-diagnosis in downstream application, confirming the possibility of precise personalized healthcare solutions.
% \ifdefined\icmlanonymous 
% \else
% Our code is available at \url{https://github.com/Raiiyf/ECGTwin}.
% \fi
\end{abstract}

\section{Introduction}

Personalized ECG generation enables the creation of a patient’s ECG digital twins across a range of cardiac conditions. It not only retains the key benefits of traditional population-level ECG generation, such as providing data for rare diseases and supporting cardiology education \citep{diffusets}, but also introduces a transformative opportunity for ECG auto-diagnosis by enabling personalized models that are finetuned on a patient’s own ECG data, allowing the model to focus on individual-specific features \citep{pecg_healthcare, pecg_review}. Since ECG signals can vary significantly among patients---even under the same cardiac condition---this personalized approach offers improved diagnostic accuracy \cite{pECGmonitor} compared to conventional models trained on population-level datasets. Together, these applications highlight personalized ECG generation as a promising direction for future research.

ECG generative model have been discussed by many work \cite{megan,text2ecg,diffusets},
but generating personalized ECG signals presents two new challenges: \textbf{(1) Extract individual features from a reference ECG without ground truth.} While deep learning models have demonstrated the ability to capture individual patient characteristics from medical data \citep{Yang2024-qt}, it remains unclear how to effectively leverage these methods for extracting personal features specifically from reference ECG signals. Unlike demographic statistics, individual features extracted from ECG signals lack ground truth for supervision, making their extraction and verification more challenging. 
Prior work on personalized ECG generation \citep{lavq} has adopted Vector Quantization \citep{VQVAE} based approach to encode individual features. However, this approach complicates training, as the feature extractor and generative model are tightly coupled and must be optimized jointly.

\textbf{(2) Inject different types of conditions into generative process.} The generation of personalized ECGs requires incorporating various types of conditional information, such as individual patient features, target cardiac diagnosis, and demographic attributes. While the abundance of conditions offers the potential for fine-grained control, it also poses a significant challenge: if not properly integrated, these conditions may confuse the generative model, leading to degraded performance or even model collapse \citep{Fetaya_understanding, forget_collapse}.
Previous work on ECG generation has been largely unconditioned or conditioned only on high-level labels, which is simple but limits both the diversity and controllability of the generated results.

\begin{figure*}[t]
    \centering
    \includegraphics[width=1.0\linewidth]{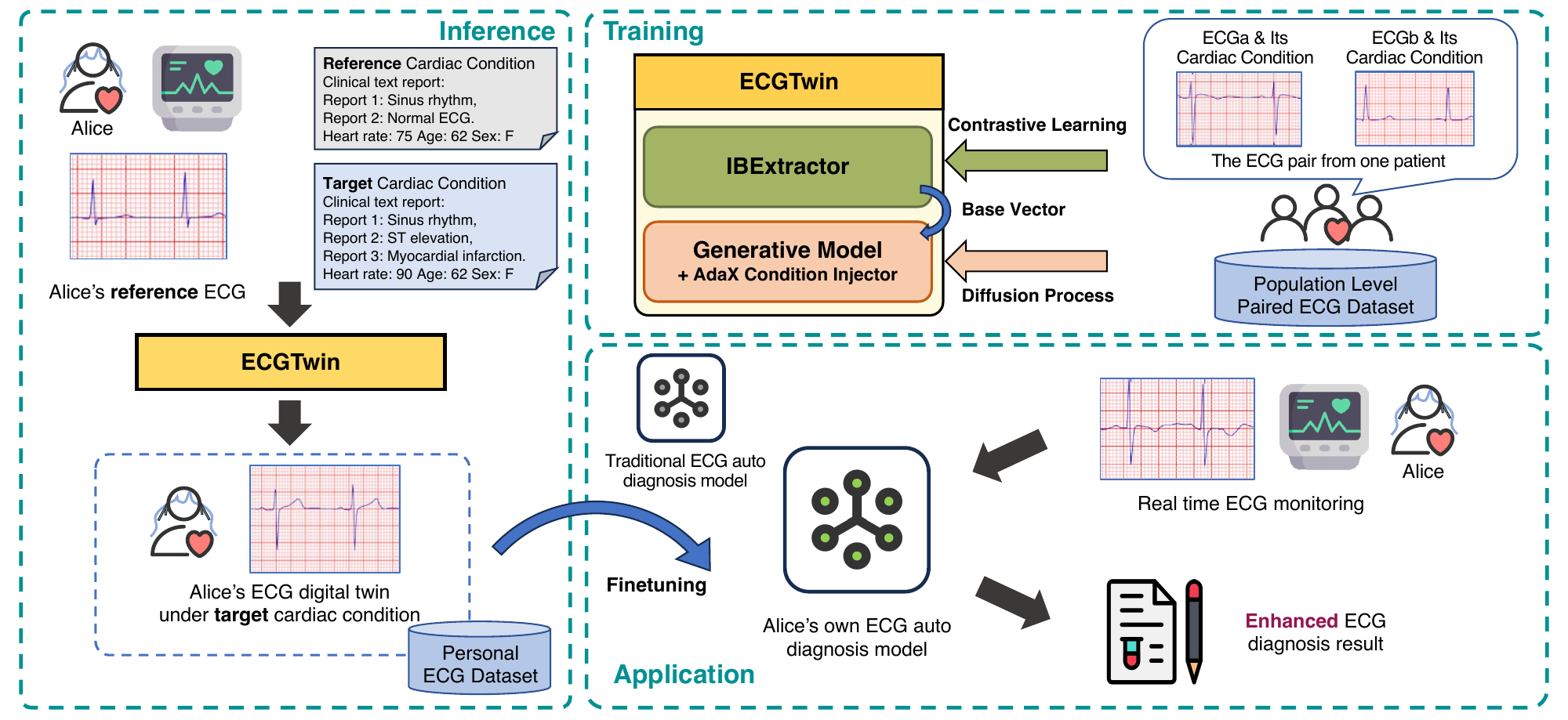}
    \caption{\textbf{Training, inference and application of \modelname.} Trained via contrastive learning and diffusion objectives respectively, and acting in a two-stage manner, our model is capable of simulating the plausible ECG signal (i.e. ECG digital twins) under a specified cardiac condition with input of a reference ECG and associated cardiac condition. The generated ECG digital twins carries more individual-relevant patterns and can be used to enhance the ECG diagnosis driven by deep learning methods.}
    \label{fig:overview}
\end{figure*}

To address aforementioned challenges, in this paper, we introduce \textbf{\modelname}, a diffusion-based model designed for personalized ECG generation. Given a reference ECG signal along with the reference and target cardiac condition, our model synthesizes high-quality ECG digital twins tailored to the target conditions (Fig.~\ref{fig:overview}). To effectively extract the base vector shared across a patient's ECGs without ground truth, we build an \textit{Individual Base Extractor} and train it separately using contrastive learning in a self-supervised manner. For the generation process, we adopt the Denoising Diffusion Probabilistic Model (DDPM) \cite{ddpm}. To better incorporate various types of conditional information into the diffusion process, we equip the noise prediction model with our proposed \textit{AdaX Condition Injector}, which enables controllable and effective conditioning through two dedicated pathways.
We build \modelname\ on a curated ECG pair dataset derived from the MIMIC-IV-ECG dataset \citep{mimicecg}, which is approximately 20 times larger than the dataset used in prior personalized ECG generation study. 

% Summary of contribution
In summary, our contributions are:
% \begin{enumerate}
%     \item[(1)] We formally define the task of personalized ECG generation and propose \modelname, a diffusion-based model dedicatedly designed to address two core challenges: capturing individual specific features and injecting diverse conditions.
%     \item[(2)] To capture personal features without explicit supervision, we develop an Individual Base Extractor trained via contrastive learning. For conditional control, we introduce the AdaX Condition Injector, which incorporates conditions through two dedicated pathways, enabling precise and effective modulation of the generation process.
%     \item[(3)] We conduct extensive experiments to evaluate the fidelity, personal consistency and explainability of the generated ECG digital twins. We also simulate a downstream ECG auto diagnosis application and demonstrate that \modelname\ significantly improves the performance of ECG auto diagnosis, highlighting its potential in personalized healthcare.
% \end{enumerate}

\begin{enumerate}
    \item[(1)] We formally define the task of personalized ECG generation and propose \textbf{\modelname}, a two-stage diffusion-based model capable of generating high-fidelity ECG digital twins.
    \item[(2)] We demonstrate that models trained via contrastive learning can effectively extract personalized features from raw ECG signals. Building on this insight, we develop the \textbf{Individual Base Extractor} as the first stage of \modelname.
    \item[(3)] we introduce the \textbf{AdaX Condition Injector}, which incorporates conditions through two dedicated pathways, enabling precise and effective modulation of the ECG generation process.
    \item[(4)] We conduct extensive experiments to evaluate the fidelity, personal consistency and explainability of the generated ECG digital twins. We also simulate a downstream ECG auto diagnosis application and demonstrate that \modelname\ significantly improves the performance of ECG auto diagnosis, highlighting its potential in personalized healthcare.
\end{enumerate}

\section{Problem Definition}

\begin{definition}
\textbf{Cardiac Condition}\\We use the term cardiac condition $\mathbf{c}$ to represent the state of the heart. It encompasses variables from multiple dimensions to provide a comprehensive characterization of cardiological and physiological situation. In our work, cardiac condition $\mathbf{c}$ consists of a set of textual clinical reports describing cardiological findings, heart rate reflecting cardiac features, age as an indicator of physiological status, and sex for hormonal factors. Notably, when data availability allows, the cardiac condition $\mathbf{c}$ can be flexibly extended to include additional variables, enabling a more complete physio-cardiac descriptor.
\end{definition}

\begin{definition}
\noindent\textbf{Personalized ECG Generation}\\Given a reference ECG $\mathrm{\mathbf{x}_{ref}}$ of a patient and its associated cardiac condition $\mathrm{\mathbf{c}_{ref}}$, the goal of personalized ECG generation is to simulate the plausible ECG signal $\mathbf{\hat{x}}$ that can reflect the patient’s cardiac state under the target cardiac condition $\mathrm{\mathbf{c}_{tar}}$. This is achieved by learning and sampling from the conditional distribution of $P(\mathbf{\hat{x}}|\mathrm{\mathbf{x}_{ref}},\mathrm{\mathbf{c}_{ref}}, \mathrm{\mathbf{c}_{tar}})$.
\end{definition}

\section{Method}

\begin{figure*}[t]
    \centering
    \includegraphics[width=1.0\linewidth]{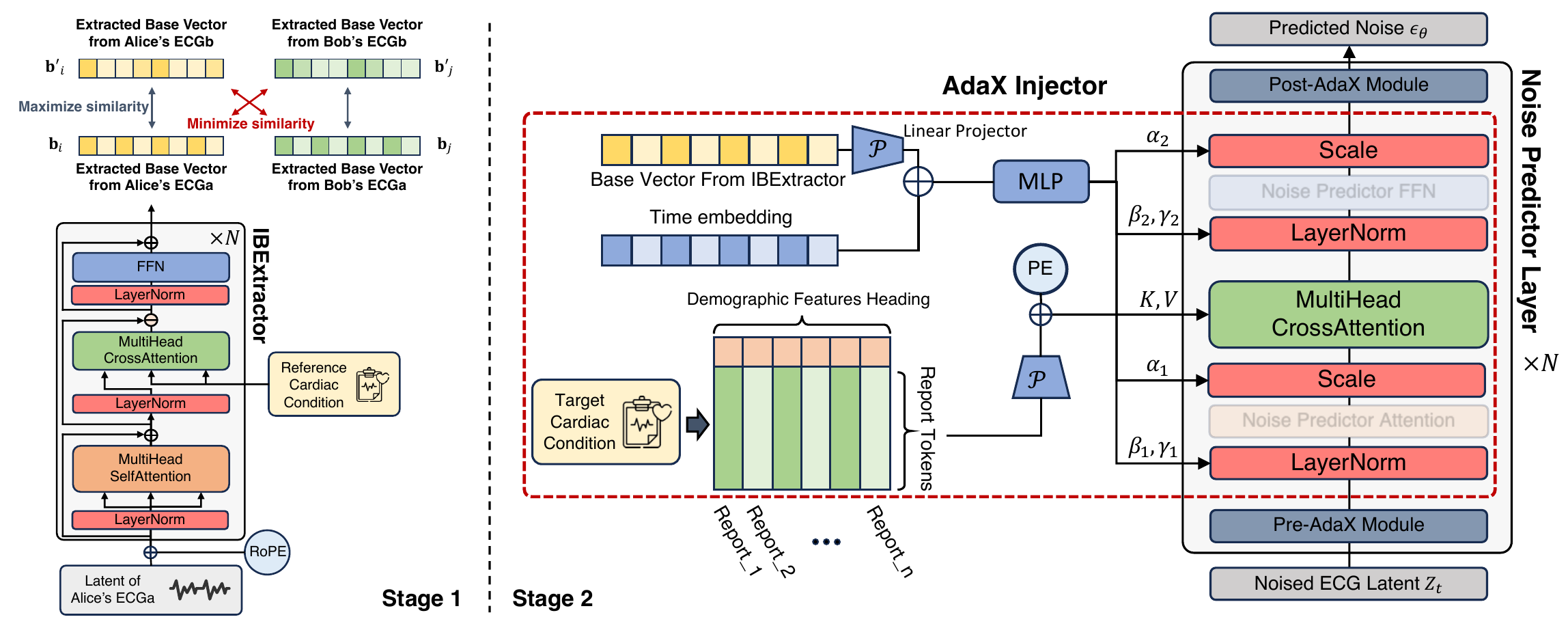}
    \caption{\textbf{Architecture of modules in \modelname's two stages.} The complete flow chart can be find in App. \ref{app:flow}.}
    \label{fig:arch}
\end{figure*}

\modelname\ performs personalized ECG generation in a two-stage manner. In first stage, the Individual Base Extractor extracts the base vector of the patient from reference ECG and its associated cardiac condition. Then, in second stage, the latent diffusion model with dedicatedly proposed AdaX Condition Injector integrates the base vector and the input target cardiac condition to generate ECG digital twins. 
% Moreover, credit to the special design in AdaX Injector, the generated result of \modelname\ can be further edited to enable a more subtle control.

\subsection{Individual Base Extractor}

Given a reference ECG $\mathrm{\mathbf{x}_{ref}}$ and its associated cardiac condition $\mathrm{\mathbf{c}_{ref}}$, we introduce an \textit{Individual Base Extractor} to extract the base vector $\mathbf{b}$ of the patient. The base vector $\mathbf{b}$ serves as a compact representation of individual-specific traits and is designed to remain invariant across ECGs recorded under different cardiac conditions for the same patient. As the substitution of reference ECG $\mathrm{\mathbf{x}_{ref}}$ and cardiac condition $\mathrm{\mathbf{c}_{ref}}$, we only forward extracted base vector $\mathbf{b}$ to the diffusion process for personalized ECG generation. This can be interpreted as a decomposition of the target conditional distribution:
$P(\mathbf{\hat{x}}|\mathrm{\mathbf{x}_{ref}},\mathrm{\mathbf{c}_{ref}}, \mathrm{\mathbf{c}_{tar}}) \rightarrow P(\mathbf{\hat{x}}|\mathbf{b}, \mathrm{\mathbf{c}_{tar}})$, where $\mathbf{b} = f_{\mathrm{IBE},\theta}(\mathrm{\mathbf{x}_{ref}},\mathrm{\mathbf{c}_{ref}})$ and $\theta$ stands for learnable parameters. 
This two-stage computation framework offers several advantages. First, it explicitly designates personalized features in the generation process, enabling the diffusion model to more effectively capture individual-specific information by reducing redundant conditioning signals. Second, it decouples the extraction of personalized information from the generative model, resulting in more stable training and simplified optimization. Additionally, we choose to extract base vector from the VAE encoded ECG latent $\mathrm{\mathbf{z}_{ref}}$ rather than directly from raw ECG signal, i.e. $\mathbf{b} = f_{\mathrm{IBE},\theta}(\mathrm{\mathbf{z}_{ref}},\mathrm{\mathbf{c}_{ref}})$, so as to better align the base vector with the latent space used by the generation model.

The remaining challenge is how to train the Individual Base Extractor to accurately learn the base vector, especially in the absence of ground-truth labels for supervision. To address this, we adopt a self-supervised learning approach. The pretext training task is to maximize the similarity of base vectors extracted from ECGs of the same patient, while minimizing the similarity of base vectors from different patients. Inspired by the CLIP Loss \citep{CLIP}, we construct a dataset of ECG pairs, where each pair consists of two ECGs recorded from the same patient. The loss function for the Individual Base Extractor is defined as:

\scalebox{0.95}{%
\begin{minipage}{\linewidth}
\begin{align}
    \mathcal{L}_{IBE} = - \frac{1}{NB}\sum_{k=1}^{N}\sum_{i=1}^{B} \frac12 
    \left( \log \frac{\exp\left( \mathrm{Sim}(\mathbf{b}_i, \mathbf{b'}_i) / {\tau} \right)}
    {\sum_{j \neq i} \exp\left( \mathrm{Sim}(\mathbf{b}_i, \mathbf{b'}_j) / {\tau} \right)} \right.
    \nonumber \\
    + \left. \log \frac{\exp\left( \mathrm{Sim}(\mathbf{b'}_i, \mathbf{b}_i) / {\tau} \right)} 
    {\sum_{j \neq i} \exp\left( \mathrm{Sim}(\mathbf{b'}_i, \mathbf{b}_j) / {\tau} \right)} \right)
\end{align}
\end{minipage}
}

\noindent where the $\mathbf{b}_i$ and $\mathbf{b'}_i$ are base vectors extracted from the i-th ECG pair. $\mathrm{Sim}(\cdot,\cdot)$ is the similarity function and we use the cosine similarity. $B$ is the batch size, $N$ is number of batches, and $\tau$ is the learnable temperature parameter. By explicitly pairing two ECGs from the same patient and viewing as positive samples, we can effectively leverage the hidden personalized features from a population level dataset as supervising signals. It is worth noting that cases where multiple pairs, e.g. $(\mathbf{b'}_i, \mathbf{b}_i)$ and $(\mathbf{b'}_j, \mathbf{b}_j)$, originate from the same patient could potentially introduce ambiguity into the model. However, since the number of ECG pairs from any patient is far less than the total number of ECG pairs, we can consider the positive matches outside designated pairings are extremely sparse, even within a large batch. As a result, the probability of such collisions is close to zero, and their impact on training can be safely ignored. An additional benefit of ECG pairing is that the same dataset can be directly used for training ECG generation model in the next stage, ensuring intrinsic consistency even when the components are optimized separately.

We implement the Individual Base Extractor function $f_{\mathrm{IBE},\theta}$ based on Transformer encoder\citep{transformer}. As shown in Fig.~\ref{fig:arch}, an cross-attention layer is added to receive the reference cardiac condition $\mathbf{c_{ref}}$. The conditional mechanism of $\mathbf{c_{ref}}$ aligns with the AdaX Cardiac Condition Pathway, which will be detailed in section~\ref{sec:adax}. 
% Intuitively, we modify the residual connection by subtracting the output of the cross-attention layer from the shortcut representation. This design encourages the model to disregard the influence of the current cardiac condition, of which the base vector should be invariant. 
Nevertheless, to enhance robustness, we randomly mask $\mathrm{\mathbf{c}_{ref}}$ at a fixed ratio during training and replace it with a special learnable embedding. This prepares the model to handle scenarios where $\mathrm{\mathbf{c}_{ref}}$ is unavailable at inference time.

\subsection{AdaX Condition Injector}
\label{sec:adax}

Given a base vector $\mathbf{b}$ encapsulating personalized features and a target cardiac condition $\mathrm{\mathbf{c}_{tar}}$ specifying desired morphological features of results, the ECG generation process should handle different kinds of conditional information meticulously so as to synthesize the patient's ECG digital twins of high quality and fidelity. In \modelname, we adopt latent diffusion process to model the target conditional distribution $P(\hat{\mathbf{z}}_0|\mathbf{b}, \mathbf{c}_\text{tar})$ by iteratively denoising the ECG latent $\hat{\mathbf{z}}_t$. During the process, a noise predictor must take the timestep $t$, the base vector $\mathbf{b}$, and the target cardiac condition $\mathrm{\mathbf{c}_{tar}}$ as conditional inputs to compute the current noise $\epsilon_\theta(\hat{\mathbf{z}}_t, t, \mathbf{b}, \mathrm{\mathbf{c}_{tar}})$ required by the DDPM reverse process. 

These conditional inputs are heterogeneous in both semantic meaning and numerical dimension, and thus should be handled through separate, specialized ways. Notably, the target cardiac condition $\mathbf{c}_{\mathrm{tar}}$ carries the most detailed information and directly governs the waveform of the generated ECG signal. Therefore, an effective and expressive mechanism for encoding cardiac condition is critical for successful controllable ECG generation. To achieve this, we design the \textit{AdaX Condition Injector} module as the conditioning interface of the noise predictor, enabling effectively integration all relevant conditions. As illustrated in Fig.~\ref{fig:arch}, the module processes conditions through two distinct pathways:

\subsubsection{Cardiac Condition Pathway} 
We utilize each component of the cardiac condition --- the clinical reports, sex, age, and heart rate --- to construct a cardiac condition sequence that serves as the key and value in a cross-attention mechanism. Note that an ECG signal is often associated with multiple clinical reports, each describing different aspects such as rhythm, morphology, or diagnosis. We treat each report as a token and employ the nomic-embed-text-v1.5 model \citep{nomic} to obtain embedding $\mathbf{e} (\in \mathbb{R}^{768})$ for each report token. Compared with traditional methods like byte-pair encoding (BPE) \citep{bpe} or concatenating all reports into a single prompt \citep{diffusets}, our report-level tokenization allow the model to selectively attend to different reports and adaptively emphasize the most relevant information. 

For remaining features, we compile sex, age, and heart rate into a vector $\mathbf{p} (\in \mathbb{R}^3$), where age and heart rate are zero-score normalized, and sex is binary encoded as 1 (male) and 0 (female). Then, the vector is duplicated and appended to each report embedding as headings, ensuring that these features are always accessible regardless of which report is being attended to. Next, the resulting augmented embeddings are stacked together to form the cardiac condition sequence. This sequence is then linearly projected into the embedding space of the noise predictor to ensure dimensional compatibility and representation consistency. Finally, positional encoding (PE) is applied to preserve the hierarchical information among the report tokens:
\begin{align}
\label{eq:cc_process}
K,V = \mathrm{Stack}(\{\text{Concat}(\mathbf{e}_i,\mathbf{p})\}^{n}_{i=1})\cdot W + \text{PE}
\end{align}
% \begin{align}
% K,V = W \cdot \bigcup (\{ \mathbf{e}_i \oplus \mathbf{p}\}^{n}_{i=1}) + \text{PE}
% \end{align}

\subsubsection{Base Vector and Time Pathway} We design the conditioning pathway for the base vector and time embedding using adaptive normalization, as both represent global information and should be injected into the model holistically. Specifically, we obtain the time embedding $\mathbf{t}$ following the sinusoidal encoding method in DiT original implementation\citep{dit}. Meanwhile, the base vector $\mathbf{b}$, extracted from the Individual Base Extractor, is linearly projected to produce a dimension-aligned embedding $\mathbf{\bar{b}}$. The two embeddings are added element-wise and passed through a Multi-Layer Perceptron (MLP) to predict the scaling factor $\alpha$, the LayerNorm shift and scale parameters $\beta$ and $\gamma$, respectively:
\begin{align}
    \alpha, \beta, \gamma =\text{MLP}(\mathbf{t} + \mathbf{\bar{b}})
\end{align}
These parameters are then used to adaptively normalize and scale the latent feature $\mathbf{z}$ as follows:
\begin{align}
    \mathbf{z}_{\text{layernorm}} = \frac{\mathbf{z}-\mathrm{E}[\mathbf{z}]}
    {\sqrt{\mathrm{Var}[\mathbf{z}]+\epsilon}} \cdot \gamma+\beta,\ 
    \mathbf{z}_{\text{scale}} = \alpha \cdot \mathbf{z}
\end{align}
where $\epsilon$ is a small constant added for numerical stability.

Another advantage of AdaX Condition Injector is its compatibility with \textit{Prompt-to-Prompt} editing\citep{prompt2prompt}, which enables a further fine-grained and flexible control over the results, crediting to the cross-attention mechanism integrated. For more details of this post-generation ECG editing technique, please refer to App.~\ref{app:p2p}. 

\subsection{Personalized ECG Generation}

We unify aforementioned Individual Base Extractor and AdaX Injector within the framework of a latent diffusion process \citep{ldm} for personalized ECG generation. In the diffusion forward process, the clean ECG latent is progressively corrupted by Gaussian noise according to:
\begin{align}
    \mathbf{z}_t = \sqrt{\bar{\alpha}_t} z_0 + \sqrt{1 - \bar{\alpha}_t} \epsilon_t \ , \epsilon_t \sim \mathcal{N}(0, \mathbf{I})
\end{align}
where $\bar{\alpha}_t$ are noise scheduling hyperparameters with formulations: $\alpha_t := 1 - \beta_t$ and $\bar{\alpha}_t := \prod_{s=1}^{t} \alpha_s$. In the diffusion reverse process, the model iteratively denoises $\mathbf{z}_t$ by sampling $\mathbf{z}_{t-1}$ from $\mathcal{N}(\mu_t, \sigma_t^2 \mathbf{I})$, where:
\begin{align}
\mu_t &:= [\sqrt{\alpha_t}(1 - \bar{\alpha}_{t-1})\mathbf{z}_t + \sqrt{\bar{\alpha}_{t-1}}(1 - \alpha_t)\hat{\mathbf{z}}_0] / (1 - \bar{\alpha}_t), 
\nonumber\\
\hat{\mathbf{z}}_0 &:= [\mathbf{z}_t - \sqrt{1 - \bar{\alpha}_t}\hat{\epsilon}_\theta(\mathbf{z}_t, t, \mathbf{b},\mathbf{c}_\text{tar})] / \sqrt{\bar{\alpha}_t}; \\
\sigma_t^2 &:= (1 - \alpha_t)(1 - \bar{\alpha}_{t-1}) / (1 - \bar{\alpha}_t)
\end{align}
Here, the noise predictor $\hat{\epsilon}_\theta$ takes the noisy latent $\mathbf{z}_t$, diffusion timestep $t$, the base vector $\mathbf{b}$, and the target cardiac condition $\mathbf{c}_\text{tar}$ as inputs. It is trained using the simplified denoising score matching loss as in DDPM:
\begin{align}
    \mathcal{L}_{\modelname} = \Vert \epsilon_t - 
    \hat{\epsilon}_\theta(\mathbf{z}_t, t, \mathbf{b},\mathbf{c}_\text{tar})\Vert^2
\end{align}
Finally, a pre-trained VAE Decoder \citep{vae} is employed to reconstruct the denoised latent $\mathbf{z}_0$ back to signal space, yielding personalized ECG digital twin $\mathbf{\hat{x}}$. Notably, the training objective can be readily extended to more recent formulations such as flow matching losses \cite{flowmatching}, and we leave it for future exploration.

\begin{table*}[tb]
\centering
\caption{\textbf{Three-level evaluation result.} Best values among each architectures are marked in \textcolor{blue}{blue}, and global-best value is bolded in \textbf{\textcolor{red}{red}}.}
\resizebox{\linewidth}{!}{
\begin{tabular}{c|l|cccc|c|c}
\toprule
 \multirow{2}{*}{\textbf{Branch}} & \multirow{2}{*}{\textbf{Model Name}} &\multicolumn{4}{|c|}{\textbf{Signal Level}} &\textbf{Feature Level} &\textbf{Diagnostic Level}
\\
  &  &\textbf{FID ($\downarrow$)} &\textbf{Precision ($\uparrow$)} &\textbf{Recall ($\uparrow$)} &\textbf{F1 ($\uparrow$)} 
 &\textbf{HR-MAE ($\downarrow$)} &\textbf{Clip ($\uparrow$)}
\\ \midrule
\multirow{2}{*}{Baseline} & DiffuSETS-p & 
\ms{245}{28.0} & \ms{0.848}{0.003} & 
\ms{0.849}{0.004} & \ms{0.849}{0.002} & 
\ms{10.76}{0.58} & \ms{0.709}{0.002}\\
 & LAVQ & \ms{234}{0.3} & \ms{0.829}{0.004} &
\ms{0.519}{0.008} & \ms{0.639}{0.006} &
\ms{38.96}{0.04} & \ms{0.674}{0.001} \\
\midrule
 & UNet-\textit{CA} &
\ms{175}{8.7} & \textbf{\textcolor{red}{\ms{0.925}{0.003}}} & 
\ms{0.755}{0.008} & \ms{0.831}{0.004} &
\textbf{\textcolor{red}{\ms{4.63}{0.30}}} & \ms{0.780}{0.001} \\
UNet & UNet-\textit{adaLN} & 
\ms{34}{1.1} & \ms{0.884}{0.004} &
\ms{0.847}{0.011} & \ms{0.865}{0.007} &
\ms{8.22}{0.66} & \ms{0.763}{0.001} \\
 & ECGTwin (\textit{Ours}) & 
 \textbf{\textcolor{red}{\ms{18}{3.1}}} & \ms{0.883}{0.003} &
 \textcolor{blue}{\ms{0.860}{0.004}} & \textcolor{blue}{\ms{0.871}{0.002}} &
 \ms{7.31}{0.56} & \textcolor{blue}{\ms{0.784}{0.001}} \\
\midrule
 & DiT-\textit{CA} & 
 \ms{99}{4.9} & \ms{0.855}{0.003} &
 \textbf{\textcolor{red}{\ms{0.903}{0.006}}} & \ms{0.879}{0.002} &
 \ms{9.04}{0.26} & \ms{0.783}{0.001} \\
DiT & DiT-\textit{adaLN} & 
\ms{51}{8.1} & \textcolor{blue}{\ms{0.873}{0.003}} &
\ms{0.893}{0.004} & \ms{0.883}{0.003} & 
\ms{7.73}{0.37} & \ms{0.752}{0.001} \\
 & ECGTwin (\textit{Ours}) &
 \textcolor{blue}{\ms{26}{2.8}} & \ms{0.872}{0.002} & 
 \ms{0.896}{0.002} & \textbf{\textcolor{red}{\ms{0.884}{0.002}}} & 
 \textcolor{blue}{\ms{7.03}{0.50}} & \textbf{\textcolor{red}{\ms{0.789}{0.000}}} \\
\bottomrule
\end{tabular}
}
\label{tab:table_1}
\end{table*}

\section{Experiment}

\subsection{Experiment Settings}

\subsubsection{Datasets}
We use the publicly available \textbf{MIMIC-IV-ECG} \citep{mimicecg} dataset for the training and evaluation of our method, and link it to MIMIC-IV-Clinical dataset \citep{mimiciv} via "subject\_id" to retrieve necessary data items. To construct paired samples for the sake of effective training, we group the ECG records by "subject\_id", and exclude patients with only a single record. For each remaining patient, we generate all possible pairs of ECG records, which results in $\binom{n_i}{2}$ samples for a patient with $n_i$ records. Following this procedure, we obtain a training set of 6,408,782 ECG record pairs and a testing set of 399,499 ECG pairs. More details related to MIMIC-IV-ECG dataset and the introduction of external validation dataset PTB-XL \citep{ptbxl} can be found in App.~\ref{app:inplement}.

\subsubsection{Implementations}
All of our trainings and tests are based on PyTorch 2.1.1, on GeForce RTX 3090, DiT based model iterates approximately 140 time steps per second when generating batch is set to 10 while Unet based model iterates approximately 60 time steps per second on the same environment.
\ifdefined\icmlanonymous
Our code is available in the \textbf{supplementary material}. 
\else
Our code is available at \url{https://github.com/Raiiyf/ECGTwin}.
\fi
For details of model implementation, please refer to App.~\ref{app:inplement}.

\subsubsection{Baselines}
We compare our method with LAVQ \cite{lavq}, which to the best of our knowledge is the only existing approach designed for personalized ECG generation. Although the task setting in LAVQ differs from ours and requires more information for generation, we adapt it for evaluation on the MIMIC-IV-ECG dataset to the extent possible. Additionally, we construct a personalized version of the population-level state-of-the-art method DiffuSETS \citep{diffusets}, denoted as DiffuSETS-p, by simply concatenating the reference ECG latent with the diffusion noise. This adaptation serves as a baseline to evaluate the effectiveness of our proposed condition injection strategy.

\subsection{Generation Quality Evaluation}

In this section, we demonstrate that \modelname\ is capable of generating high-quality ECG digital twins. We use the three-level evaluation protocol \citep{diffusets} to comprehensively assess the generation results from the scope of data distribution (using FID, Precision, Recall and F1) to semantic alignment (using HR-MAE and CLIP Score). We build and test \modelname\ on two types of noise predictor backbone architectures: DiT and Unet. To highlight the effectiveness and adaptivity of our proposed AdaX Condition Injector, we perform ablation studies on both backbones by limiting the conditioning mechanism to a single pathway. For clarity of the core mechanism, models incorporating the Cardiac Condition Path are suffixed by CA (cross-attention) and models with the Base Vector and Time Path are suffixed by adaLN (adaptive LayerNorm). Detailed descriptions of the evaluation metrics and the ablated model variants are provided in App.~\ref{app:quality_eval}. The test result are summarized in Tab.~\ref{tab:table_1}. 

\textbf{Compared with other ECG digital twin generation methods, the two variants of \modelname\ present the most outstanding overall performance.} This superiority can be attributed to our two-stage framework, which effectively reduces confounding information and simplifies the training process. Furthermore, both \modelname\ models incorporating AdaX Condition Injector outperform their respective ablated counterparts in each branches, highlighting the its effectiveness in integrating conditions through two dedicated paths. In particular, models equipped with the Cardiac Condition Path consistently achieve higher scores in clinical text report alignment (measured by CLIP Score). This observation directly reflects the synergy between our report-level tokenization strategy and the cross-attention mechanism --- an appropriate match for the structured nature of ECG clinical text reports. In summary, this three-level evaluation confirms that \modelname\ can generate ECG of high fidelity and diversity, thus is also compatible for applications of conventional (i.e. population-level) ECG synthesis models. Regarding the possible distribution bias brought by single training set, we also perform an \textbf{external validation on PTB-XL dataset for the generalization ability evaluation}, please refer to App.~\ref{app:quality_eval} for details.

\subsection{Personal Consistency Assessment}
\label{sec:consistency}
In this section, we first test the efficacy of our Individual Base Extractor (IBExtractor) trained in a self-supervised manner. Then we show that the ECG digital twins generated by \modelname\ encapsulate personalized features of the reference ECG within the scope defined by the base vector.

\subsubsection{Efficacy of Individual Base Extractor}
\begin{figure*}[tb]
    \centering
    \includegraphics[width=1.03\linewidth]{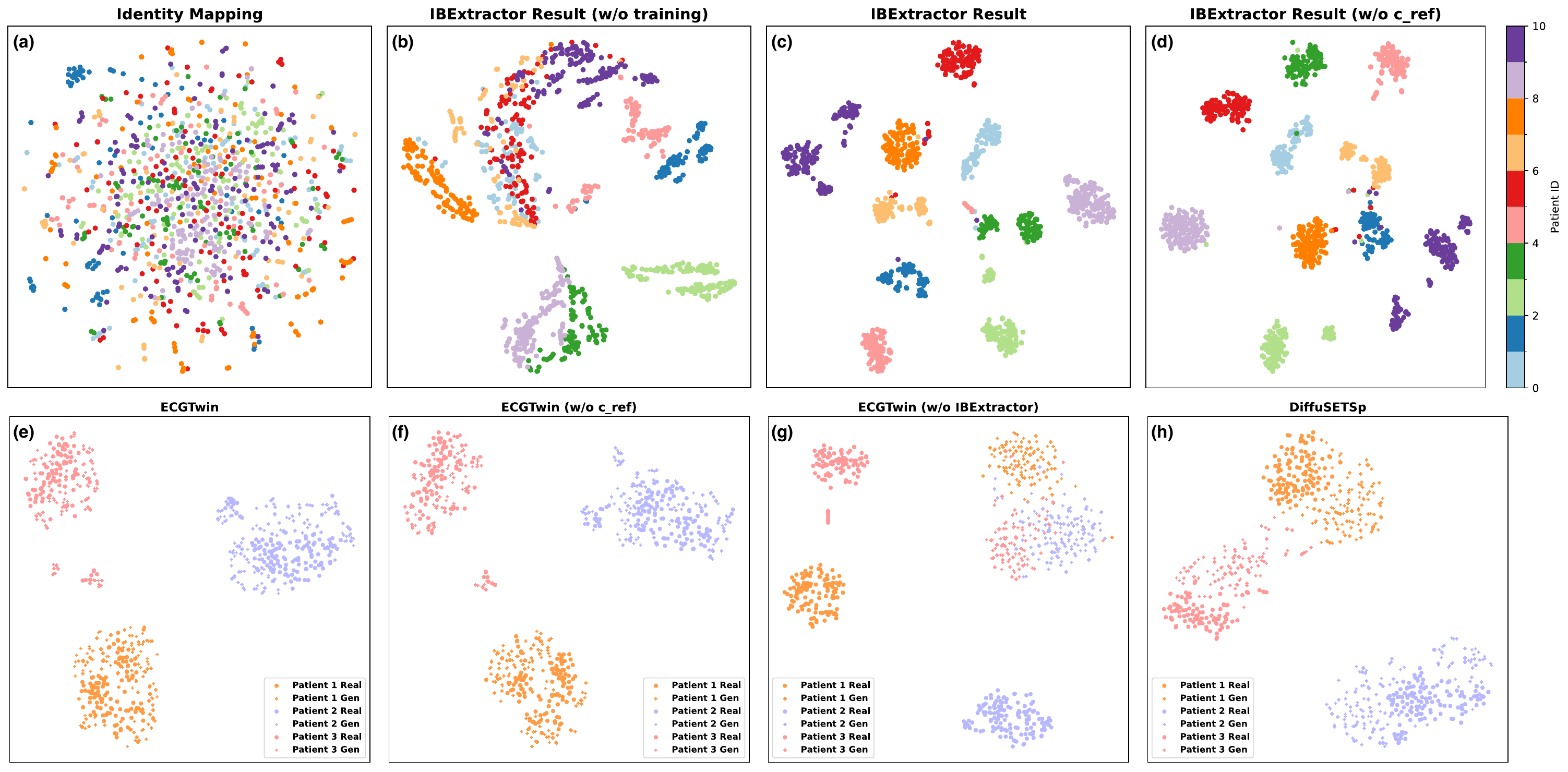}
    \caption{\textbf{The visualization result of base vector t-SNE embeddings.} (a)--(d): Base vectors of real ECGs from ten patients; (e)--(h): Base vectors of real ECGs and generated digital twins from three patients}
    \label{fig:ibe_tSNE}
\end{figure*}

We choose ten patients from the test set, each with more than 85 ECG records. From these records, we obtain a set of base vectors $B$, where each element is derived using the Individual Base Extractor. To show the original data distribution as a reference, we also test the identity mapping, i.e. directly flattening the ECG latent as output.
Fig.~\ref{fig:ibe_tSNE} presents the t-SNE visualization of extracted base vectors. To numerically assess the intra- and inter-individual similarity, we propose a \textbf{similarity score} s: 

\scalebox{0.9}{%
\begin{minipage}{\linewidth}
\begin{align}
    \label{eq:sim_score}
    s = \frac{1}{|B|}\sum_{\mathrm{b}_i\in B} 
    \left(
    \frac{1}{|B_I|}\sum_{\mathrm{b}_j \in B_I}
    \frac{\mathrm{b}_i\cdot \mathrm{b}_j}
    {\Vert \mathrm{b}_i\Vert \Vert \mathrm{b}_j\Vert}
    % \text{Sim}(\mathrm{b}_i, \mathrm{b}_j)
    % \right.\nonumber
     - 
    % \left.
    % \frac{1}{|B-B_I|}\sum_{\mathrm{b}_j \notin B_I}
    \frac{1}{|B^c_I|}\sum_{\mathrm{b}_j \notin B_I}
    \frac{\mathrm{b}_i\cdot \mathrm{b}_j}
    {\Vert \mathrm{b}_i\Vert \Vert \mathrm{b}_j\Vert}
    % \text{Sim}(\mathrm{b}_i, \mathrm{b}_j)
    \right)
\end{align}
\end{minipage}
}

\noindent where $B_I$ is the $i$-th patient's base vector set, and $B^c_I$ is the complement of $B_I$, i.e. $B\setminus  B_I$. This score can reflect the overall base vectors quality from the cognateness for the same individual and the distinction among different individuals. We also use the traditional silhouette coefficient \citep{silhouette} with euclidean distance metric to quantify clustering quality in the embedded space. The quantitative results are presented in Tab.~\ref{tab:consistency}.

\begin{table}[tb]
\centering
\caption{\textbf{Similarity score and silhouette coefficient result.} Best values are bolded, while the second-best are underlined.}
\resizebox{\columnwidth}{!}{
\begin{tabular}{@{}l|cc@{}}
\toprule
 \textbf{Methods} & \textbf{Similarity Score} & \textbf{Silhouette Coefficient} \\
\midrule
\multicolumn{3}{c}{Efficacy of Individual Base Extractor (Fig.~\ref{fig:ibe_tSNE}(a)--(d))}\\
\midrule
Identity Mapping      & 0.0022 & -0.1447 \\
IBExtractor w/o training      & 0.0894 & 0.2298  \\
IBExtractor          & \underline{0.2808} & \underline{0.6439}  \\
IBExtractor w/o c\_ref        & \textbf{0.3079} & \textbf{0.6738} \\
\midrule
\multicolumn{3}{c}{Generation Consistency (Fig.~\ref{fig:ibe_tSNE}(e)--(h))}\\
\midrule
ECGTwin        & \underline{0.3334} & \textbf{0.7401} \\
ECGTwin w/o c\_ref  & \textbf{0.3440} & \underline{0.7380} \\
ECGTwin w/o IBExtractor   & 0.1689 & 0.1715 \\
DiffuSETp & 0.2412 & 0.6577\\
\bottomrule
\end{tabular}
}
\label{tab:consistency}
\end{table}

From the comparison of Fig.~\ref{fig:ibe_tSNE}(a) and (c), we observe that \textbf{the base vectors extracted by Individual Base Extractor form distinct clusters for each patient}, while the original latents exhibit a stochastic distribution. This entropy reduction illustrates the model’s ability to capture personalized ECG characteristics. Fig.~\ref{fig:ibe_tSNE}(b) surprisingly shows immature clustering, which may be affected by reference cardiac condition $\mathbf{c}_\text{ref}$. Nevertheless, compared with Fig.~\ref{fig:ibe_tSNE}(c), the substantial difference brought by model training highlights the value of our self-supervised training strategy. We also test the performance of our method in the absence of $\mathbf{c}_\text{ref}$, and it even show superior result on these ten patients according to the numerical assessment in Tab.~\ref{tab:consistency}, which confirms the model's robustness originating from our masked training. 
We conduct a further analysis of Individual Base Extractor regarding to $\mathbf{c}_\text{ref}$, which is provided in App.~\ref{app:ibe_scale}. \textbf{Furthermore, to demonstrate the robustness of Individual Base Extractor, we test its resistant ability in App.~\ref{app:ibe_noise} when the input ECG is corrupted by noise at different scales.}

\subsubsection{Generation Consistency}
Having validated the pre-trained Individual Base Extractor, we now use it to assess personalization consistency in ECG generation. Specifically, we select three patients from test set and generates their ECG digital twins. We then analysis the base vector extracted from both the real and generated ECGs together by t-SNE visualization and numerical evaluation (Fig.~\ref{fig:ibe_tSNE}(e)--(h) and Tab.~\ref{tab:consistency}). We include two baselines for comparison: the result when the base vector is zero out and the result of DiffuSET-p.

Our findings show that whether with the auxiliary of $\mathbf{c}_\text{ref}$, \textbf{\modelname\ can generate ECG accurately preserving patient-specific features} (Fig.~\ref{fig:ibe_tSNE}(e) and (f)). In contrast, when the base vector from stage 1 is omitted (Fig.~\ref{fig:ibe_tSNE}(g)), the generated ECGs lose individualized characteristics, resulting in base vectors that no longer cluster meaningfully. This confirms the necessity of the base vector and the efficacy of the Individual Base Extractor in Stage 1. Finally, in both visual and quantitative terms, the personal patterns retained by DiffuSETSp (Fig.~\ref{fig:ibe_tSNE}(h)) are less distinct than those captured by our proposed method. This supports the effectiveness of our two-stage framework, and again underscores the critical role of the AdaX Condition Injector in conditional generation.

\subsection{Personalized Healthcare Simulation}

In this section, we simulate a downstream application of personalized ECG diagnosis and show that the ECG digital twins generated by \modelname\ can further enhance the performance of ECG auto diagnosis model through data augmentation. To prevent data leakage, we select 293 patients from the \modelname\ test dataset, each of whom has more than 10 ECG records, for this diagnostic evaluation. A baseline population-level ECG diagnosis model based on ResNet \citep{resnet} is first trained on the remaining ECG records in test set. This model is tasked with binary classification to distinguish between normal and abnormal ECG signals. For the personalized setting, we designate the earliest ECG of each patient as the reference and generate multiple ECG digital twins using predefined, patient-agnostic cardiac conditions as targets. These target conditions are chosen to include a broad range of abnormalities and ensure that the generated dataset is balanced between normal and abnormal classes. The baseline model is then fine-tuned individually for each patient using their own set of ECG digital twins, resulting in personalized diagnosis models. To further validate the effectiveness of \modelname, we compare it against models augmented by ECG synthesized in population-level scope. We report the evaluation results for these 293 patients in Tab.~\ref{tab:ecg_diagnosis} from both patient wise (computing metrics for each patient then averaging) and population wise (directly averaging among all test data).

\begin{table}[tb]
\centering
\caption{\textbf{ECG auto diagnosis test.} \modelname\ use DiT architecture as noise predictor backbone. Best values are bolded.}
\resizebox{\columnwidth}{!}{
\begin{tabular}{@{}c|c|cc|cc@{}}
\toprule
\multirow{2}{*}{\textbf{Method Scope}} & \multirow{2}{*}{\textbf{Augmented By}} & \multicolumn{2}{|c|}{\textbf{Patient-Wise}} & \multicolumn{2}{c}{\textbf{Population-Wise}} \\
 &  & \textbf{Acc.} & \textbf{Macro-F1} & \textbf{Acc.} & \textbf{Macro-F1}  \\
\midrule
\multirow{2}{*}{\textbf{Population}} & N/A (Base Model) &0.755	&0.672	&0.759	&0.753  \\
 & DiffuSETS & 0.797	&0.696	&0.800	&0.785  \\
 \midrule
\multirow{3}{*}{\textbf{Personalized}} & DiffuSETS-p 
&0.806	&0.694	&0.809	&0.783  \\
& LAVQ &0.810 &0.727 &0.808 &0.795  \\
& \modelname\ (\textit{Ours}) & \textbf{0.816}	&\textbf{0.731}	&\textbf{0.819}	&\textbf{0.804} \\
\bottomrule
\end{tabular}
}
\label{tab:ecg_diagnosis}
\end{table}

The result of simulation experiments \textbf{validate the potential of personalized healthcare and show that \modelname\ is well-suited for this promising application.}
Compared with population-level diagnosis model, the personalized diagnosis model augmented with \modelname\ generated ECG digital twins demonstrates substantial performance improvements. This can be attributed to the personal features-preserving nature of our method, which enables the finetuned model to focus on individual-specific patterns, leading to more targeted and accurate diagnosis. Furthermore, among all personalized augmentation methods, \modelname\ yields the best performance. This result highlights its superior ability to capture the most relevant and distinct personal features via contrastive learning, while generating high-quality ECG signals through effective condition injection. 

\subsection{Case Study and Interpretability Analysis}

\begin{figure}[tb]
    \centering
    \includegraphics[width=1.0\linewidth]{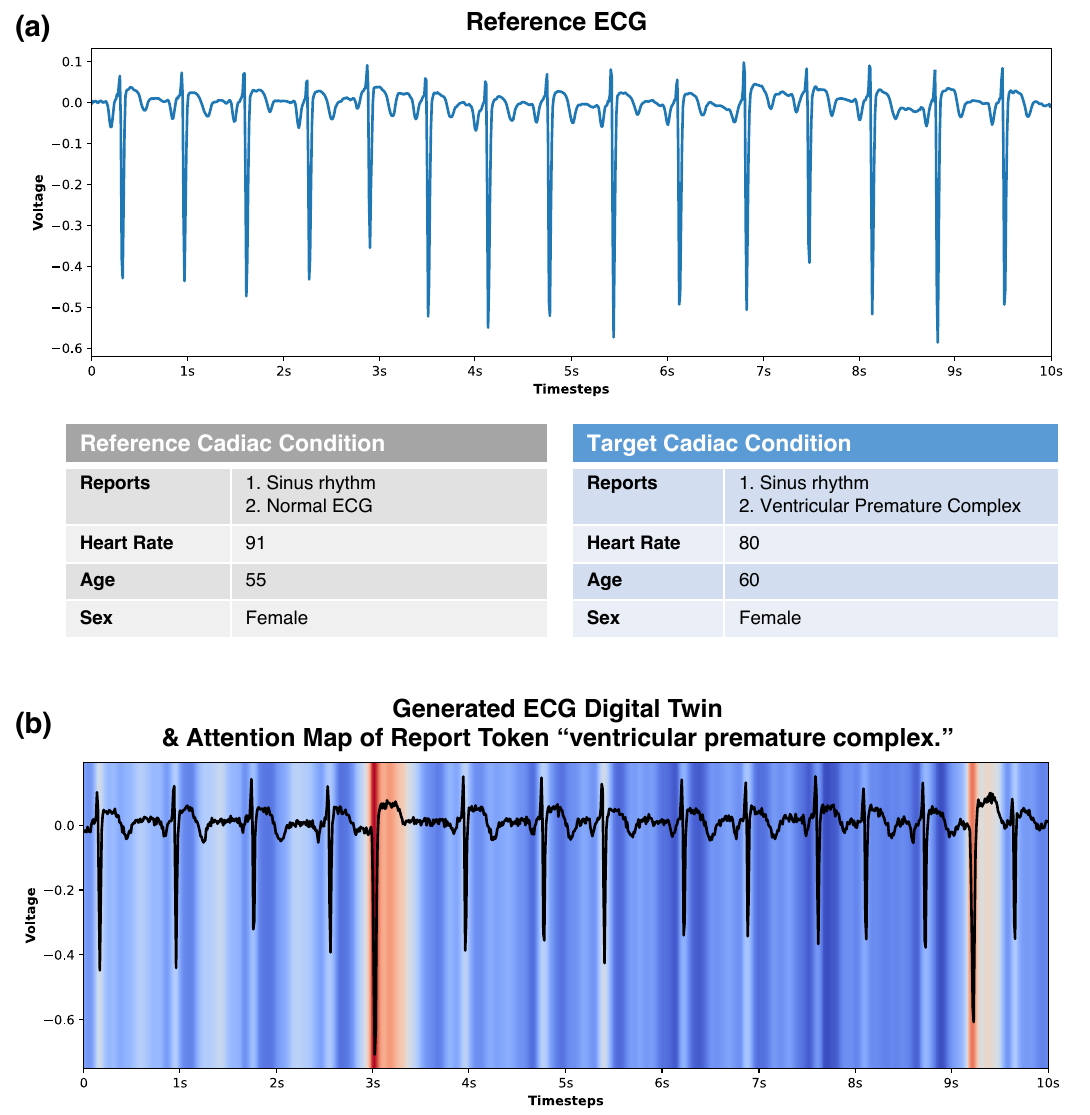}
    \caption{\textbf{Case study of personalized ECG generation.} \\(a): Input data, including the reference ECG, associated cardiac condition, and the target cardiac condition. (b): The ECG digital twin generated by \modelname, along with the average cross-attention map of the report token PVC. Redder regions indicate higher amount of attention.}
    \label{fig:case_attn}
\end{figure}

In this section, we take the common cardiac event \textit{"Ventricular Premature Complex (PVC)"} as a representative case to illustrate the process of personalized ECG generation and the intrinsic explainability of \modelname. 
\textit{PVC} has a specific appearance of the QRS complexes and T waves on ECG, which are different from normal readings. To simulate this, we generate a ECG digital twin with \textit{PVC} features of one patient, using her normal ECG as reference. 
Notably, to establish a direct correlation between attention map and raw ECG waveform, we perform the generation on a special version of \modelname\ that operates directly in the signal space without a VAE. 
Fig.~\ref{fig:case_attn}(a) shows the complete input setup of personalized ECG generation, including the reference ECG signal, its associated cardiac condition (providing auxiliary description of current state), and the target cardiac condition (specifying the desired cardiac state). The generation result is presented in Fig.~\ref{fig:case_attn}(b). For clarity, we only display ECG lead V1 here. For holistic view of all 12-leads and \textbf{additional case studies}, please refer to App.~\ref{app:case}.

Next, we visualize the average cross-attention map of report token \textit{PVC} in the \textit{AdaX Condition Injector} at diffusion timestep $T=100$. As shown in Fig.~\ref{fig:case_attn}(b), malformed QRS complexes, which is exactly the characteristic pattern of \textit{PVC}, appear at the localizations where the report token \textit{PVC} places most of its attention. This alignment demonstrates that \modelname\ is capable of correctly identifying and localizing clinically meaningful patterns during personalized ECG generation. Such behavior not only enhances the interpretability of the model, which is an essential requirement for clinical adoption \citep{clinical_explainablitiy}, but also lays the foundation for a fine-grained post-generation editing, as we further explore and showcase in App.~\ref{app:p2p}.

\section{Conclusion}

In this paper, we present \modelname, a novel two-stage diffusion-based model for personalized ECG generation that addresses the challenges of individual feature extraction and multi-condition integration. By combining a contrastively trained \textit{Individual Base Extractor} with the diffusion process equipped by \textit{AdaX Condition Injector}, our method enables controllable and interpretable generation of ECG digital twins that faithfully retain personal characteristics. Through comprehensive experiments, we demonstrate that \modelname\ achieves superior performance in generation quality, while also exhibiting strong downstream utility. These results suggest that \modelname\ is well-positioned to advance the development of personalized healthcare and foster broader adoption of generative models in clinical applications.

\section*{Impact Statement}

This paper presents work whose goal is to advance the field of Machine
Learning. There are many potential societal consequences of our work, none
which we feel must be specifically highlighted here.

\bibliography{ECGTwin}
\bibliographystyle{icml2026}

%%%%%%%%%%%%%%%%%%%%%%%%%%%%%%%%%%%%%%%%%%%%%%%%%%%%%%%%%%%%%%%%%%%%%%%%%%%%%%%
%%%%%%%%%%%%%%%%%%%%%%%%%%%%%%%%%%%%%%%%%%%%%%%%%%%%%%%%%%%%%%%%%%%%%%%%%%%%%%%
% APPENDIX
%%%%%%%%%%%%%%%%%%%%%%%%%%%%%%%%%%%%%%%%%%%%%%%%%%%%%%%%%%%%%%%%%%%%%%%%%%%%%%%
%%%%%%%%%%%%%%%%%%%%%%%%%%%%%%%%%%%%%%%%%%%%%%%%%%%%%%%%%%%%%%%%%%%%%%%%%%%%%%%
\newpage
\appendix
\onecolumn
% \addcontentsline{toc}{part}{Appendix}

\section{Related Work}

\subsection{ECG Generative Methods}

ECG generation methods have evolved significantly over the years, progressing from early unconditional approaches or those conditioned on high-level labels \citep{megan, SSSDECG} to more refined control using textual clinical reports \citep{text2ecg, diffusets}. Some researchers \citep{lavq} have explored personalized ECG generation using Generative Adversarial Networks (GANs) \citep{gan} combined with Vector Quantization (VQ) based methods. In their framework, a VQ-variational autoencoder (VAE) \citep{VQVAE} is used for feature disentanglement and should be optimized jointly with GAN components, which would complicate the already delicate training process typical of GANs. Furthermore, their approach requires an additional input ECG reflecting the target cardiac condition as part of the conditioning signal, which may be impractical when the condition is complex or rare. Beyond GANs, researchers have applied more expressive architectures, such as autoregressive models \cite{text2ecg} and diffusion models \cite{SSSDECG, diffusets}, to conditional ECG generation. However, their works do not incorporate a reference ECG as input and therefore cannot achieve personalized generation. In our work, we use the powerful diffusion model for ECG generation and build an Individual Base Extractor to obtain the personalized ECG base vector, which is trained separately using contrastive learning, thus will not add extra burden to the optimizing of diffusion process.

\subsection{Conditional Generation}

Methods for incorporating conditional information into generative models can generally be categorized into three types \citep{dit}:
(1) In-context conditioning. The condition signals are directly appended to the input, either along the sequence length axis or the embedding (channel) dimension.
(2) Adaptive normalization. The learnable parameters in normalization layers---such as the scale and shift parameters $\gamma$ and $\beta$ in LayerNorm \citep{layernorm}---are dynamically modulated by the condition signals.
(3) Cross-attention. The model architecture is augmented with additional multi-head cross-attention layers \citep{transformer}, where the condition signals are used as the Key and Value inputs, allowing the model to attend selectively to the conditioning information during generation.

Prior work \citep{dit} has conducted extensive experiments demonstrating the superiority of zero-initialized adaptive LayerNorm in terms of both generated image fidelity and computational efficiency. However, other researchers \citep{prompt2prompt} have also shown that cross-attention layers offer more nuanced control, particularly due to the interpretability of the attention maps, which capture meaningful relationships between the input and the conditional signals. Moreover, a common defect for these methods is that they could not handle complex, heterogeneous conditions. Therefore, in our work, we develop the AdaX Condition Injector, combining the benefit of both adaptive LayerNorm
and cross-attention and processing different types of conditions in separate pathways, to effectively integrate all the conditions concerning with the personalized generation process. 

\newpage
\section{\modelname\ Framework} 
\label{app:flow}

\begin{figure}[ht]
    \centering
    \includegraphics[width=1.0\columnwidth]{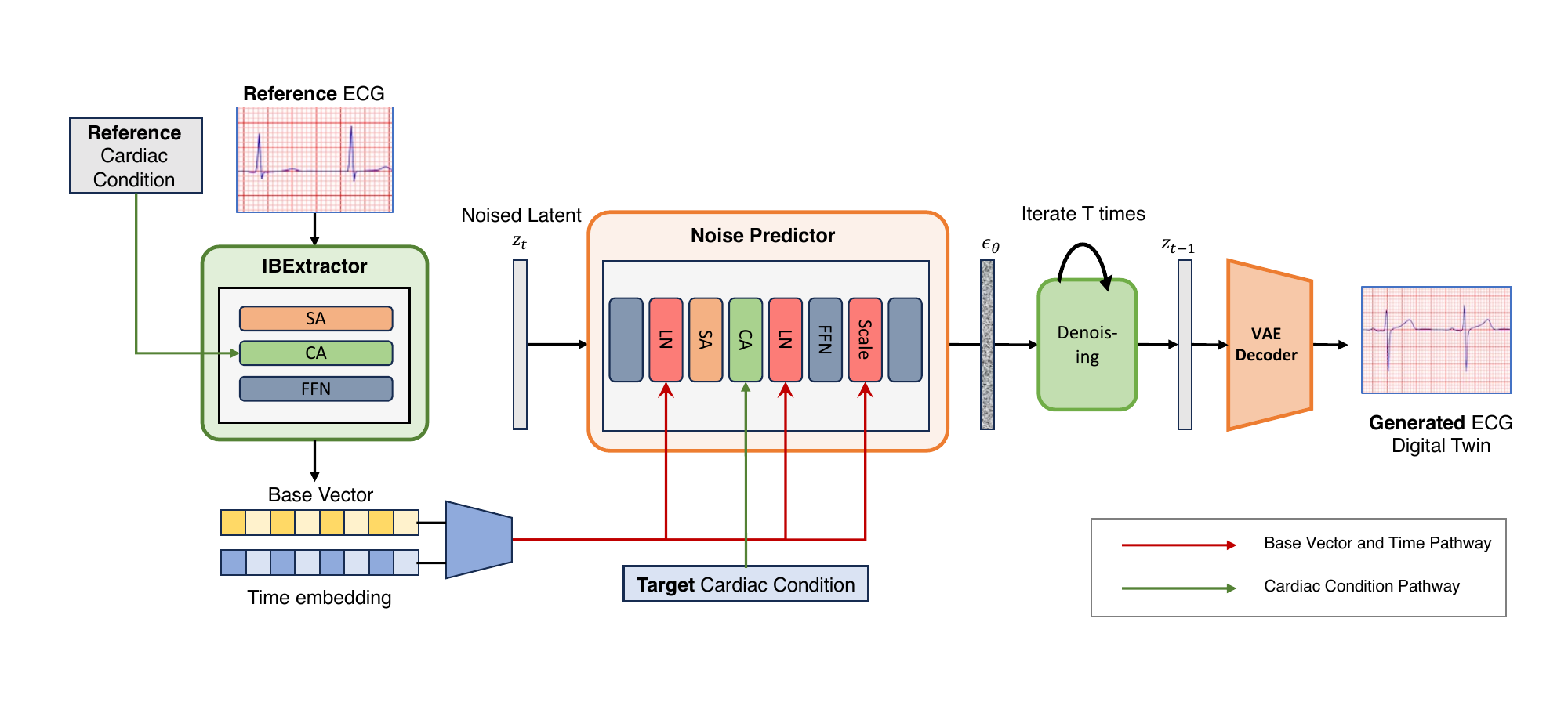}
    \caption{\textbf{The flow chart of \modelname.} The Individual Base Extractor first extracts the base vector of the patient from reference ECG and reference cardiac condition, then the latent diffusion model with AdaX Condition Injector integrates the base vector, current diffusion timestep and target cardiac condition to generate ECG digital twins by repeatedly denoising a latent sampled from Gaussian distribution.}
\end{figure}
\clearpage

\newpage
\section{Experiment details} 
\label{app:inplement}
\subsection{Dataset}
\subsubsection{MIMIC-IV-ECG}
We use the publicly available \textbf{MIMIC-IV-ECG} dataset \citep{mimicecg} for the training and evaluation of our method. MIMIC-IV-ECG contains 800,035 records from 159,538 unique patients, where each record comprising a 10-second 500Hz 12-lead ECG waveform and 1 to 17 associated clinical text reports written in English. First we downsample the ECG waveform to 102.4Hz, resulting a ECG signal matrix $\in \mathbb{R}^{1024\times12}$. We then link the MIMIC-IV-ECG dataset to MIMIC-IV-Clinical dataset \citep{mimiciv, physionet} via subject\_id to retrieve the ECG owner's age and sex, and compute the heart rate using recorded RR interval. When RR interval shows anomaly (i.e. 0 or 65536 ms), we directly parse the ECG waveform by utilizing the XQRS detector from WFDB toolkit\citep{sharma2023wfdb} to manually obtain QRS interval then compute the heart rate. After preprocessing, we retain 794,372 ECG records with complete and valid information. These are split into 744,372 records for preparation of training dataset and 50,000 records for preparation of testing dataset.

\begin{table}[ht]
\centering
% \resizebox{\columnwidth}{!}{
\begin{tabular}{c|cc|c}
\toprule
 \multirow{2}{*}{\textbf{Dataset}} & \multicolumn{2}{|c|}{\textbf{MIMIC-IV-ECG}} & \multirow{2}{*}{\textbf{PTB-XL}} \\
 & \textbf{Train split} & \textbf{Test split} & \\
\midrule
\# of ECG Records      & 744,372 & 50,000 & 21799\\
\# of ECG Pairs      & 6,408,782 & 399,499 & 4269 \\
Avg. \# of ECG Records Per Patient    & 4.98 & 4.89 & 1.16 \\
Max \# of ECG Records Per Patient & 260 & 168 & 10\\
\bottomrule
\end{tabular}
% }
\caption{\textbf{Statistics of MIMIC IV ECG and PTB-XL Dataset}}
\label{tab:dataset_stat}
\end{table}

\begin{figure}[hb]
    \centering
    \includegraphics[width=0.9\columnwidth]{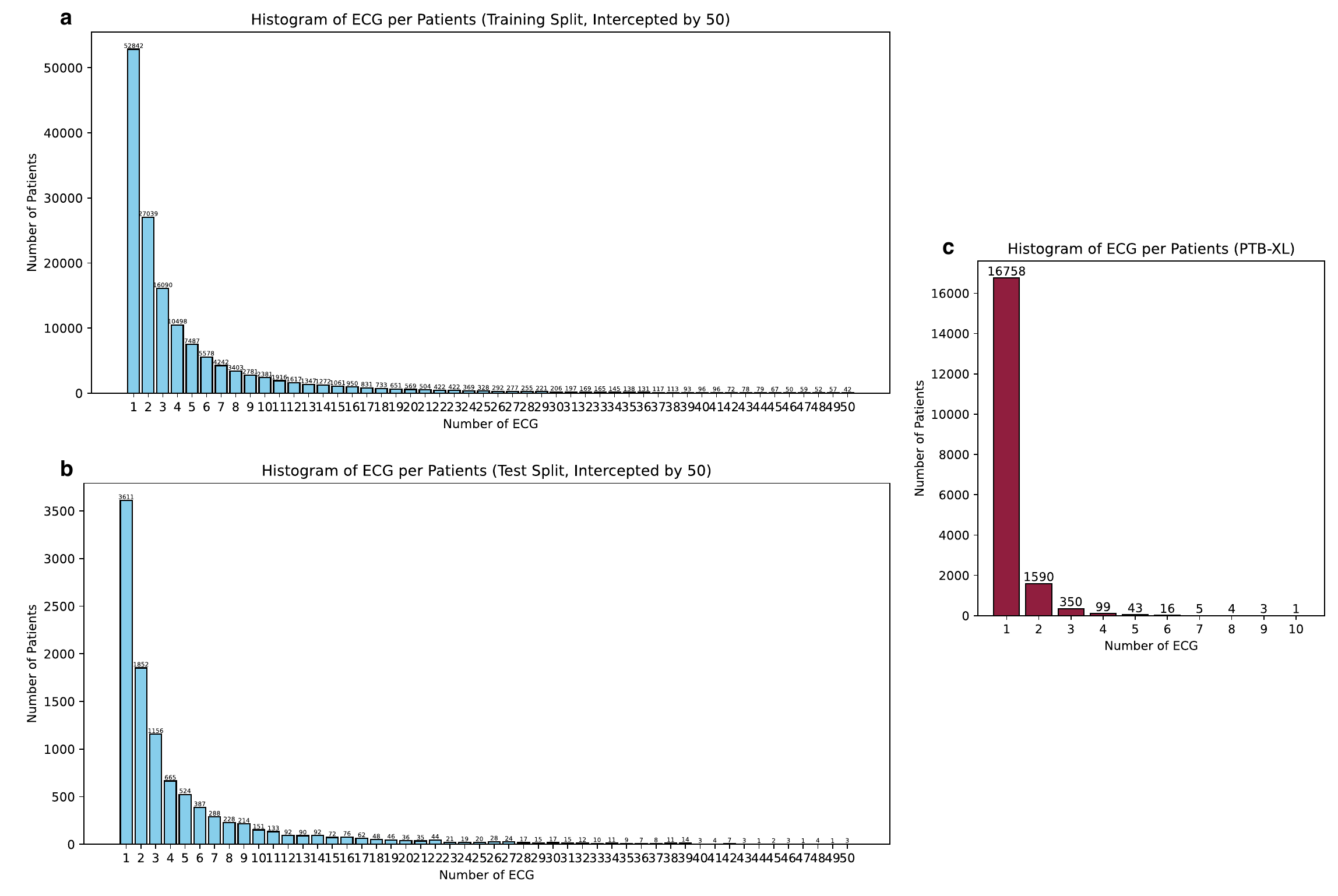}
    \caption{\textbf{The histogram of per patient ECG records number.} (a) and (b): MIMIC-IV-ECG dataset; (c): PTB-XL dataset.}
    \label{fig:dataset_hist}
\end{figure}

To construct paired samples for training the Individual Base Extractor and DDPM noise predictor, we group the ECG records by subject\_id, and exclude patients with only a single record. The distribution and statistics of ECG counts per patient are shown in Table~\ref{tab:dataset_stat} and Figure~\ref{fig:dataset_hist}. For each remaining patient, we generate all possible pairs of ECG records, ordered chronologically to reflect potential temporal causality in the conditional distribution $P(\mathbf{\hat{x}}|\mathrm{\mathbf{x}_{ref}},\mathrm{\mathbf{c}_{ref}}, \mathrm{\mathbf{c}_{tar}})$.
This results in $\binom{n_i}{2}=\frac12n(n-1)$ samples for a patient with $n_i$ records. Following this procedure, we obtain a training set of 6,408,782 ECG record pairs and a testing set of 399,499 ECG pairs based on the aforementioned split.

\subsubsection{PTB-XL} We use PTB-XL dataset \citep{ptbxl} for external validation. It contains 21,799 12-lead ECG records from 18,885 individuals. We use the waveform data in 10-second 500Hz format and downsample them into 102.4Hz. Each record is associated with a single clinical text report, written either in English or German, along with 1--3 labels that describe the record from diagnostic, form and rhythm perspectives. To enhance the textual condition, we expand each label into a descriptive phrase and treat these as additional text reports. Consequently, after preprocessing, each record is accompanied by 2–4 clinical reports. Notably, since the PTB-XL labels do not include heart rate related values, we compute the heart rate directly from the waveform data. The pairing strategy for constructing ECG pairs in PTB-XL follows the same procedure as used for the MIMIC-IV-ECG dataset. All the ECG pairs are used for test without further splitting. The distribution and statistics of ECG record counts per patient are also shown in Table~\ref{tab:dataset_stat} and Figure~\ref{fig:dataset_hist}.

\subsection{Implementation details}
For Individual Base Extractor at stage 1, the layer of Transformer encoder is set to 3 and the model dimension, i.e. the resulting base vector dimension, is set to 256. We use a large batch size of 65,536 and use the gradient accumulation trick with mini batch of 512 for compatibility to our CUDA memory. Moreover, as stated previously, we randomly substitute the $\mathbf{c}_{ref}$ with a learnable embedding at probability of 0.15 to enhance the robustness of model. We use adamW with learning rate of 1e-3 for optimization and set number of training epochs to 40.

For the latent diffusion model at stage 2, we test both prevailing style for noise predictor, Unet \citep{unet} and DiT \citep{dit}. The training details of these two architectures are listed in Table~\ref{tab:modelpara}. We adopt the pre-trained VAE model with latent space of $\mathbb{R}^{4\times128}$ from DiffuSETS \citep{diffusets} without further parameter-tuning. The number of time step $T$ in training phase is set to 1000, and the noise $\beta_t$ of diffusion forward process is assigned to linear intervals of $[8.5\times10^{-4}, 1.2\times10^{-2}]$. Also, we randomly mask the base vector from stage 1 with zero vector at a ratio of 0.15 to retain the model's ability for non-personalized generation.

Among all hyperparameters, we observe that the learning rate has the most significant impact on generation quality. The noise predictor model fails to converge when the learning rate is set either too high or too low, and we perform a grid search to identify the best setting for stable training and good performance. For other hyperparameters, such as diffusion noise schedules, we follow configurations from previous works \citep{diffusets, dit}. We acknowledge that more intensive hyperparameter tuning could further improve \modelname’s performance. However, in this paper, our primary goal is to introduce the method and demonstrate its efficacy. Hence this additional optimization is left to future work for the research community interested in ECG digital twin generation.

All of our trainings and tests are implemented with PyTorch 2.1.1, on GeForce RTX 3090. During inference phase, DiT based model iterates approximately 140 time steps per second when generating batch is set to 10 while Unet based model iterates approximately 60 time steps per second on the same environment. Therefore, without explicitly specifying, the \modelname\ refers to the DiT version model.

\begin{table}[h!]
\centering
% \resizebox{\columnwidth}{!}{
\begin{tabular}{l|c|c|c|c}
\toprule
 \textbf{Methods} & \multicolumn{2}{c|}{\textbf{DiT}} & \multicolumn{2}{c}{\textbf{UNet}} \\
\midrule
\# Epochs      & \multicolumn{2}{c|}{30} & \multicolumn{2}{c}{30} \\
Batch Size      & \multicolumn{2}{c|}{512} & \multicolumn{2}{c}{1024}  \\
Learning Rate    & \multicolumn{2}{c|}{1e-4} & \multicolumn{2}{c}{5e-4}  \\
Model Size & \multicolumn{2}{c|}{$\sim$9M} & \multicolumn{2}{c}{$\sim$8M} \\
\midrule
Model-  & Model Dim. & 256 & kernel size & 7 \\
Specific & \# Layers & 7 & \# Levels & 6 \\
\bottomrule
\end{tabular}
% }
\caption{\textbf{Hyperparameters of different noise predictor backbones.}}
\label{tab:modelpara}
\end{table}

\newpage
\section{Supplement of Quality Evaluation}
\label{app:quality_eval}

\subsection{Three-level Evaluation}
We employ the three-level evaluation \citep{diffusets} for the generated ECG twin. The signal level focuses on the fidelity and stability of the generated signals by evaluating the distribution similarity (Fréchet Inception Distance Score, FID) and the structural resemblance between real and generated ECG signals (The improved Precision, Recall and F1 \citep{precision_recall}). The feature level examines whether the ECG signals generated by the model align with the input descriptions of patient-specific information. In this case, we test the Mean Absolute Error (MAE) between the Heart Rate extracted from the generated ECG and the heart rate specified in target cardiac condition (HR-MAE). Last but not least, the diagnostic level use CLIP scores \citep{clipscore} to assess the semantic alignment between the generated ECG signals and the clinical text reports in target cardiac condition. All the tests are repeated 5 times with the mean and standard deviation of results computed.

\subsection{Ablated Models}
To provide a penetrating analysis of our proposed AdaX Condition Injector, we design two type of condition injectors, each inherits one of its dedicated condition pathways. The model with suffix \textit{CA} preserves the report-level tokenization trick and the cross attention mechanism for cardiac condition injection while simply adding the time embedding and base vector to the noisy latent representation after necessary linear projection for dimension adaptation. The model with suffix \textit{adaLN} uses adaptive LayerNorm to integrate all the conditions. In this case, we concatenate all the reports together and obtain one text embedding. Successive process is the same as the Equation~\ref{eq:cc_process} except that the condition sequence reduces into a single vector. The resulting cardiac condition vector alongside the time embedding and base vector are finally summed together and forwarded into an MLP for modulating the factors of layernorm and scaling gate.

\subsection{External Validation}
To evaluate the generalization capability of \modelname, we perform the three-level evaluation on an external dataset, PTB-XL, without any additional fine-tuning. A detailed introduction to PTB-XL is provided in Appendix~\ref{app:inplement}. Notably, the most significant difference between the MIMIC-IV-ECG and PTB-XL datasets lies in the distribution and language of clinical text reports associated with each ECG record. As shown in Table~\ref{tab:external}, \modelname\ maintains strong performance on this out-of-distribution dataset, whereas the ablated models exhibit substantial performance degradation. This performance gap underscores the challenge of effectively injecting diverse condition types and further validates the necessity and synergy of the two-pathway design in the AdaX Condition Injector. It is also worth noting that CLIP Score metrics on the external validation set are slightly lower than those in Table~\ref{tab:table_1} across models trained on MIMIC-IV-ECG dataset. This may be attributed to the fact that many clinical text reports in PTB-XL are written in German, a language for which the employed text embedding model may not be fully optimized. Nevertheless, \modelname\ achieves the best semantic alignment performance among all baselines and retains competitive absolute scores, further confirming its robustness and adaptability.

\begin{table*}[htb]
\centering
\resizebox{\linewidth}{!}{
\begin{tabular}{c|cccc|c|c}
\toprule
 \multirow{2}{*}{\textbf{Model Name}} &\multicolumn{4}{|c|}{\textbf{Signal Level}} &\textbf{Feature Level} &\textbf{Diagnostic Level}
\\
  &\textbf{FID ($\downarrow$)} &\textbf{Precision ($\uparrow$)} &\textbf{Recall ($\uparrow$)} &\textbf{F1 ($\uparrow$)} 
 &\textbf{HR-MAE ($\downarrow$)} &\textbf{Clip ($\uparrow$)}
\\ \midrule
  DiffuSETSp & \ms{335}{69} & \ms{0.690}{0.009} & 
  \ms{0.917}{0.012} & \ms{0.784}{0.006} & 
  \ms{13.76}{0.66} & \ms{0.689}{0.003} \\
  LAVQ & \ms{588}{02} & \textbf{\ms{0.844}{0.004}} &
  \ms{0.310}{0.023} & \ms{0.455}{0.022} & 
  \ms{39.82}{0.03} & \ms{0.718}{0.002} \\
  DiT-\textit{CA} & \ms{416}{34} & \ms{0.706}{0.008} & 
  \textbf{\ms{0.950}{0.008}} & \underline{\ms{0.810}{0.007}} &
  \underline{\ms{7.35}{0.34}} & \underline{\ms{0.723}{0.002}} \\
  DiT-\textit{adaLN} & \underline{\ms{199}{26}} & \ms{0.686}{0.013} &
  \ms{0.948}{0.005} & \ms{0.796}{0.009} & 
  \textbf{\ms{6.88}{0.46}} & \ms{0.708}{0.002} \\
  ECGTwin-\textit{DiT}(\textit{Ours}) & \textbf{\ms{41}{13}} & \underline{\ms{0.743}{0.009}} & 
  \underline{\ms{0.949}{0.008}} & \textbf{\ms{0.833}{0.007}} & 
  \ms{10.23}{0.59} & \textbf{\ms{0.729}{0.002}}\\
\bottomrule
\end{tabular}
}
\caption{\textbf{Three-level evaluation result on PTB-XL dataset.} Best values are bolded, while the second-best are underlined.}
\label{tab:external}
\end{table*}

\subsection{Generated Distribution Visualization}
\begin{figure}[ht]
    \centering
    \includegraphics[width=0.4\columnwidth]{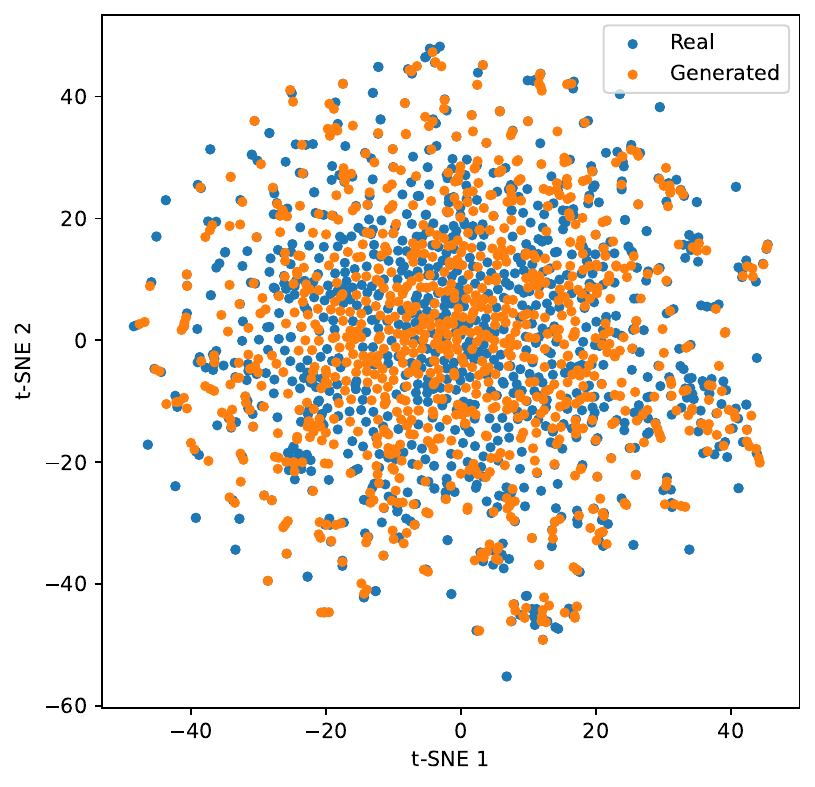}
    \caption{\textbf{The scatters of generated ECG digital twin latents and real target ECG latents.}}
    \label{fig:latents}
\end{figure}
We randomly select 1,000 ECG pairs from the test dataset and generate one ECG digital twin for each pair, conditioned on the reference ECG latent, reference cardiac condition, and target cardiac condition. We then visualize the real target ECG latents and the latents of the generated ECG digital twins using t-SNE. As shown in Figure~\ref{fig:latents}, the distribution of the generated latents closely aligns with that of the real ECG latents, with generated samples evenly dispersed among the real ones. This result provides strong evidence that \modelname\ is capable of producing realistic ECG signals that resemble real data from a latent space perspective.

\subsection{Heart Rate Scatters}
\begin{figure}[ht]
    \centering
    \includegraphics[width=0.7\columnwidth]{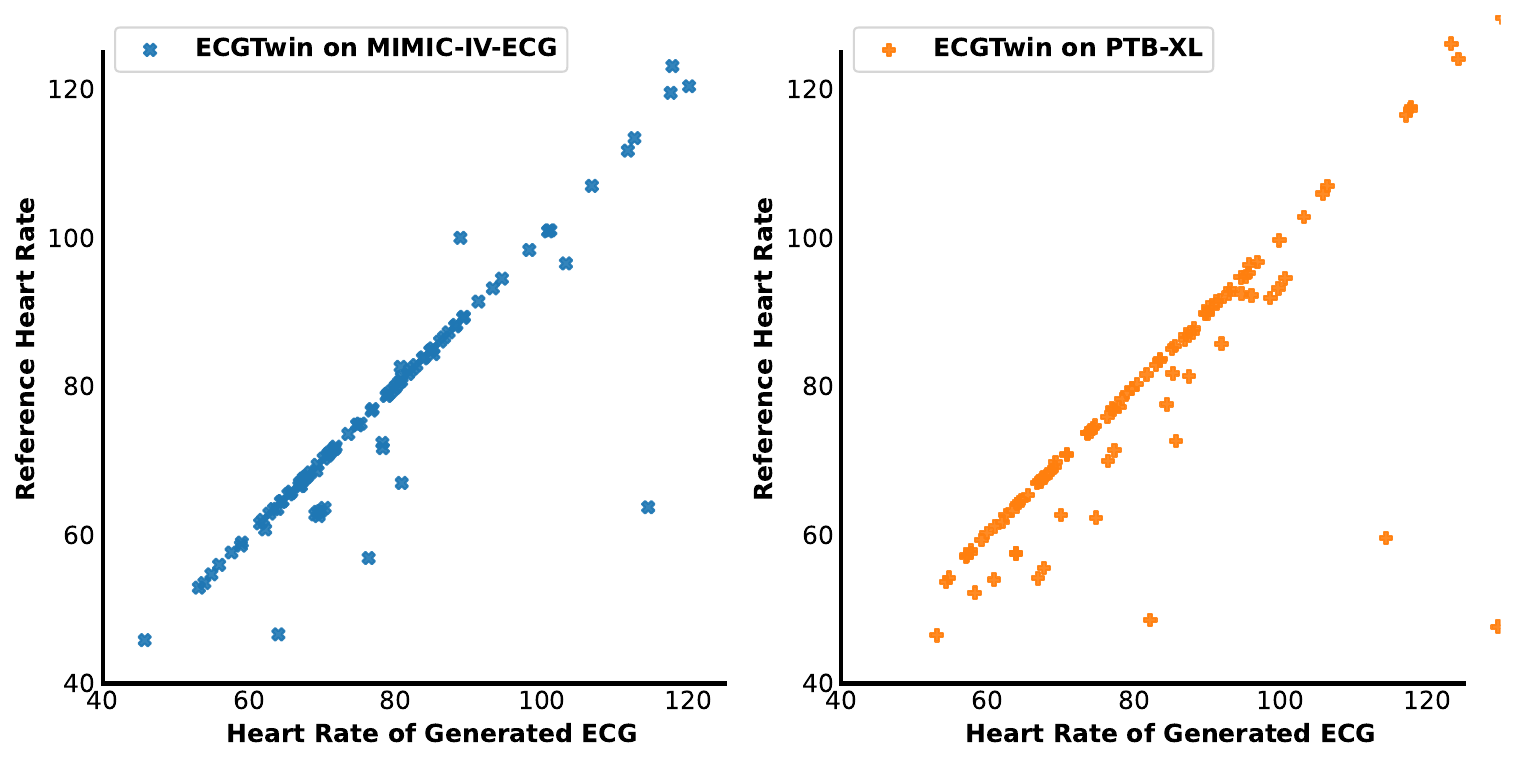}
    \caption{\textbf{The scatters of heart rate in generated ECG and input target cardiac condition.}}
    \label{fig:scatters}
\end{figure}

To evaluate whether \modelname\ can generate ECG digital twins that accurately reflect the input heart rate, we visualize scatter plots of generated versus target heart rates in Figure~\ref{fig:scatters}. Across both datasets, the data points closely align with the identity line ($y = x$), indicating that \modelname\ effectively interprets the heart rate specified in the input target cardiac condition and generates ECG digital twins that adhere to this rhythm constraint. While \modelname\ may not achieve the lowest HR-MAE among all methods, the deviation is minor and does not impair its ability to model the intrinsic relationship between input heart rate value and waveform periodicity. This confirms the model’s competence in generating temporally accurate and physiologically plausible ECG signals.

\clearpage
\newpage
\section{Supplementary Analysis of Individual Base Extractor}
\label{app:ibe_scale}

\subsection{Model Input Analysis}

In this part, we focus on the interesting observation in Table~\ref{tab:consistency}, that is, the Individual Base Extractor shows a better result when reference cardiac condition $\mathbf{c}_\text{ref}$ is absent. We try to figure out whether it is because possible bias from $\mathbf{c}_\text{ref}$ or it is just a coincidence related to specific test patient group. Note that we have verified in the main text that the model can capture personalized ECG characteristics in both cases, hence this analysis is for deeper exploration and is never intended to provide an arena for performance duelling. After all, they are the same model with the same weights but in different working status.

\subsubsection{Discussion on Reference Cardiac Condition}

We first justify the reference cardiac condition $\mathbf{c}_\text{ref}$ used in Individual Base Extractor. The intention of $\mathbf{c}_\text{ref}$ is to explicitly designate the spurious factor of which the base vector should be invariant. However, concern may be raised based on the potential data bias: in the real dataset, the ECG records of the same patient may tend to have similar cardiac conditions because the sampling is unevenly conducted on certain in-hospital period, during which the cardiac condition change is not substantial regarding to the person's lifelong time. This data bias may lead model to learn a wrongly shortcut by relying on $\mathbf{c}_\text{ref}$, which is entirely opposite to the original purpose. 

In spite of the potential bias in the training data, here we show that our model is not affected by it from test time observations. We examine the reference cardiac conditions of the ten patients used in Section~\ref{sec:consistency} efficacy experiment. We find that they not only differ intra-patient to some extent, but also show overlapping inter-patients. This implies that the distributions of $\mathbf{c}_\text{ref}$ from patient trajectories is more dispersed and intersected than expectation, which may hugely ease the intrinsic bias. Figure~\ref{fig:ibe_tSNE}(b) further confirms our assumption, since the entangled clustering result somehow reflects the distribution of $\mathbf{c}_\text{ref}$. However, in Figure~\ref{fig:ibe_tSNE}(c), points form clusters with distinct edge after training. This disentanglement indeed demonstrate that model can avoid the data bias, and thus we can tell that the performance gap may not come from data bias introduced by $\mathbf{c}_\text{ref}$.

\subsubsection{Scaling Up Test}

\begin{figure}[h!]
    \centering
    \includegraphics[width=0.4\columnwidth]{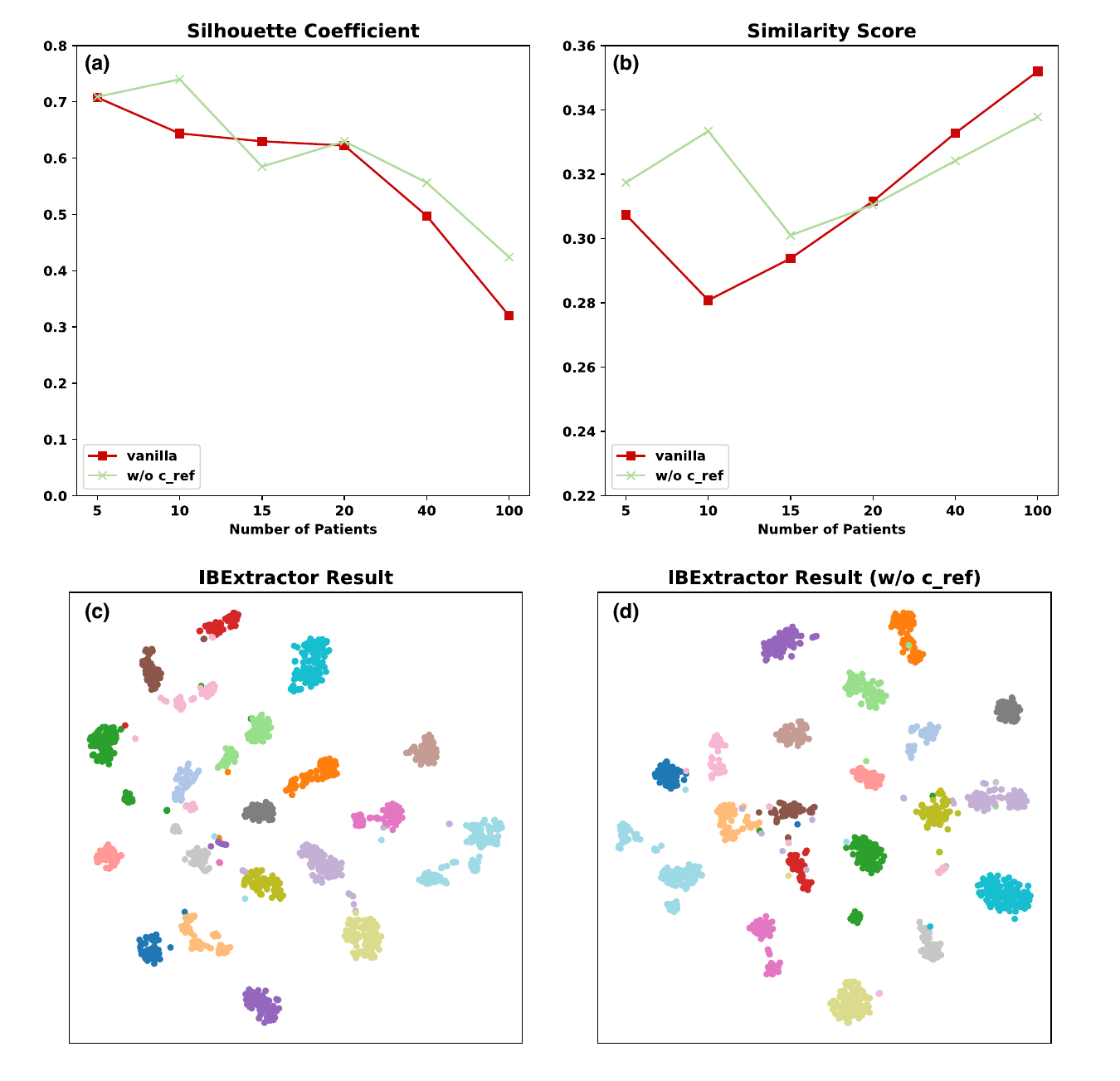}
    \caption{\textbf{The scaling up test on Individual Base Extractor with and without reference cardiac condition.} (a) and (b): Quantitative result; (c) and (d): t-SNE visualization of ECG base vector from twenty patients.}
    \label{fig:ibe_scale}
\end{figure}

We then provide a comprehensive evaluation of the performance under two input cases by selecting different numbers of patient and repeating the efficacy experiment in Section~\ref{sec:consistency}. As illustrated in Figure~\ref{fig:ibe_scale}(a) and (b), when the test group scales up, both model can capture individual-specific patterns with respect to the intra-individual similarity and inter-individual dis-similarity. Specifically, Figure~\ref{fig:ibe_scale}(c) and (d) show the t-SNE visualization of the 20 patient result, which are both impressive considering the task difficulty. From this extended assessment, we can observe that the model \textbf{with} $\mathbf{c}_\text{ref}$ shows preferable according to similarity score while the model \textbf{without} $\mathbf{c}_\text{ref}$ performs better on the scope of cluster silhouette. From this observation, we can \textbf{conclude that each method has its own merits, and considering the design of personalized ECG generation task, we choose to include $\mathbf{c}_\text{ref}$ as long as it is applicable.}

\subsection{Model Robustness Evaluation}
\label{app:ibe_noise}

\begin{figure}[t!]
    \centering
    \includegraphics[width=\columnwidth]{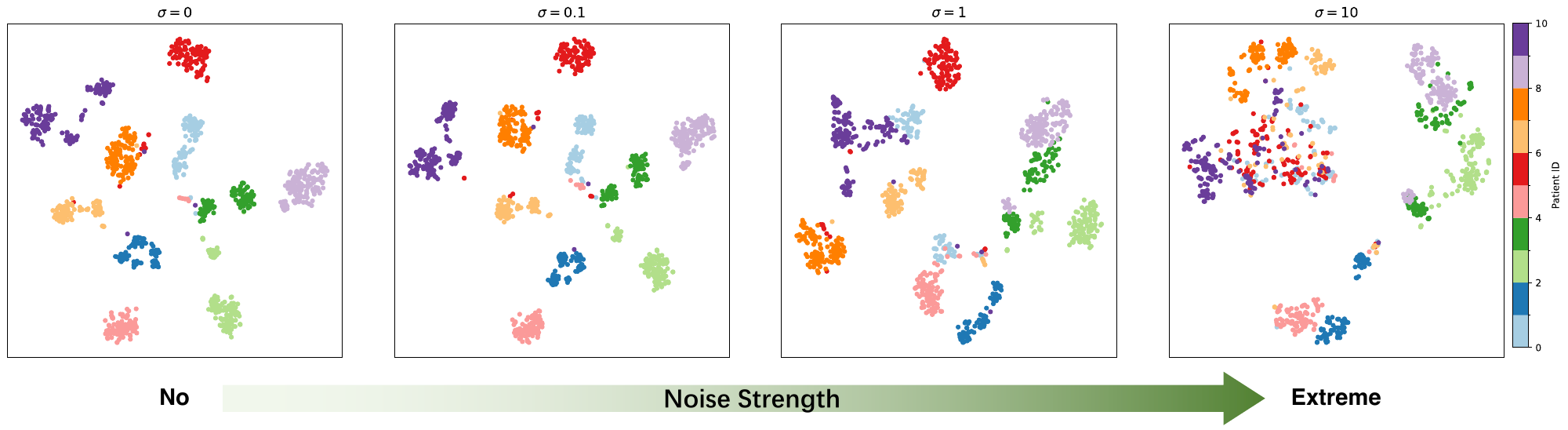}
    \caption{\textbf{Visualization of base vector clustering under different noise level.} $\sigma=0$ (no noise), $\sigma=0.1$ (mild noise), $\sigma=1$ (strong noise), and $\sigma=10$ (extreme noise).}
    \label{fig:ibe_noise}
\end{figure}

We evaluate the robustness of the Individual Base Extractor by assessing its resistance to noise when the reference ECG $\mathbf{x}{\mathrm{ref}}$ is corrupted at different noise levels. Specifically, we simulate common signal degradations encountered in clinical practice, such as electrical interference and device noise, by adding Gaussian noise with varying standard deviation $\sigma$ directly to $\mathbf{x}{\mathrm{ref}}$ in the signal space. We then visualize the clustering behavior of the extracted base vectors $\mathbf{b}$, following the setup in Section~\ref{sec:consistency}. Considering the typical voltage amplitude of ECG signals, we examine four noise regimes: $\sigma=0$ (no noise), $\sigma=0.1$ (mild noise), $\sigma=1$ (strong noise), and $\sigma=10$ (extreme noise).

As shown in Fig.~\ref{fig:ibe_noise}, the Individual Base Extractor remains largely unaffected under mild noise ($\sigma=0.1$) and continues to capture distinct patient-specific features even under strong noise conditions ($\sigma=1$). Under extreme noise ($\sigma=10$), although some feature overlap across patients emerges, the relative similarity inter patient groups are preserved. 
Overall, these results demonstrate that \textbf{the Individual Base Extractor is robust to substantial input corruption}. Moreover, because the feature extraction stage is decoupled from the generation stage, \textbf{this robustness further contributes to the stability and quality of the ECG digital twins produced by \modelname.}

\newpage
\section{Prompt-to-Prompt ECG Editing}
\label{app:p2p}

In this section, we provide technical details and experimental results regarding prompt-to-prompt editing \citep{prompt2prompt} in \modelname. This editing mechanism offers enhanced controllability over the generation process by allowing modifications to the generated results through natural language descriptions. Importantly, this functionality can be seamlessly integrated into \modelname, thanks to the Cardiac Condition Pathway in the AdaX Injector and the use of report-level tokenization. These design choices enable precise and interpretable conditioning based on textual prompts (report tokens in our case), facilitating more intuitive and flexible ECG manipulation after the initial generation.

\subsection{Theory of Prompt-to-Prompt ECG Editing}

The theoretical basis of prompt-to-prompt editing lies in the directional behavior and interpretability of cross-attention mechanisms. Typically, the attention map $M$ is defined as:
\begin{align}
\label{eq:attn_map}
    M = \text{Softmax}\left(\frac{QK^T}{\sqrt{d}}\right)
\end{align}
where $d$ is the model dimension. According to this definition, the entry $M_{ij}$ represents the attention weight of the $j$-th report token on the $i$-th latent representation. As the cross-attention operation in AdaX Injector Cardiac Condition Path determines how report tokens (i.e., prompts) influence the generated ECG, we can manipulate the generation result by injecting the attention maps $M$ that were obtained during the generation with the original prompt $\mathcal{P}$, into a second generation with the modified prompt $\mathcal{P^*}$. This allows the synthesis of an edited ECG signal $\mathbf{x^*}$ that both reflects the changes introduced by the new prompt and preserves the structural characteristics of the originally generated ECG $\mathbf{x}$.

The algorithm of prompt-to-prompt ECG editing is detailed in Algorithm~\ref{algo:p2p}. We denote by $DM(z_t,P,t,s)$ the computation of a single step $t$ of the diffusion process, which outputs the noisy latent $z_{t-1}$, and the attention map $M_t$ (omitted if not used). We also define by $M(z_t, \mathcal{P}, t, s) \{ M \leftarrow \widehat{M} \}$ the diffusion step where we override the attention map $\widehat{M}$ with an additional given map $M$, but keep the values $V$ from the supplied prompt. Furthermore, we define $Edit(M_t,M^*t ,t)$ to be a general edit function, receiving as input the $t$-th attention maps of the original and edited images during their generation. 
The specific edit function used in our work are modeled in Equation~\ref{eq:edit_func}. Specifically, to add a new report token, the attention map of new report token $M^*_t$ is appended to the source attention map $M_t$ along the token axis after a predefined diffusion timestep $\tau$:
\begin{align}
\label{eq:edit_func}
    Edit(M_t,M^*_t,t) := \begin{cases} 
\text{Concat}(M_t,M^*_t)  &\text{if } t < \tau \\
M_t  &\text{if } t \geq \tau
\end{cases}
\end{align}

\begin{algorithm}[htb]
\caption{Prompt-to-Prompt ECG editing}
\label{algo:p2p}
\begin{algorithmic}[1]
\REQUIRE A source prompt $\mathcal{P}$, a target prompt $\mathcal{P}^*$, and a random seed $s$.
\ENSURE A source ECG latent $z_0$ and an edited image $z_0^*$.
\STATE $z_T \sim \mathcal{N}(0, I)$ a unit Gaussian random variable with random seed $s$;
\STATE $z_T^* \leftarrow z_T$;
\FOR{$t = T, T-1, \ldots, 1$}
    \STATE $z_{t-1}, M_t \leftarrow DM(z_t, \mathcal{P}, t, s)$;
    \STATE $M_t^* \leftarrow DM(z_t^*, \mathcal{P}^*, t, s)$;
    \STATE $\widehat{M}_t \leftarrow Edit(M_t, M_t^*, t)$;
    \STATE $z_{t-1}^* \leftarrow DM(z_t^*, \mathcal{P}^*, t, s_t) \{ M \leftarrow \widehat{M}_t \}$;
\ENDFOR
\STATE \textbf{Return} $(z_0, z_0^*)$
\end{algorithmic}
\end{algorithm}

This technique allows a modulation to the \modelname-generated ECG to reflect the changes introduced by the new report token while preserving the overall structural and personal characteristics. Notably, we employ this method to edit the generated ECG, enabling more nuanced control over the ECG generation process. However, this approach can also be applied to edit real ECG signals with the help of DDIM reverse. This opens up the possibility for an alternative method for personalized ECG generation, which we plan to explore in future work.

% \subsection{Attention Map Visualization: More Cases}
% In this section, we present additional examples (Figure~\ref{fig:attn_map_1} and \ref{fig:attn_map_2}) of cross-attention maps from the Cardiac Condition Pathway in \modelname. Notably, the visualized heatmaps reveal that the examined report tokens exert substantial influence on clinically relevant waveform segments during the generation process. These consistent attention patterns provide further evidence that \modelname\ can effectively associate textual cardiac conditions with their corresponding morphological features in the ECG signal, thereby enhancing interpretability and clinical reliability.

% \begin{figure}[ht]
%     \centering
%     \includegraphics[width=0.9\columnwidth]{Appendix/attn_map_1.pdf}
%     \caption{More examples of cross-attention maps.}
%     \label{fig:attn_map_1}
% \end{figure}

% \begin{figure}[ht]
%     \centering
%     \includegraphics[width=0.9\columnwidth]{Appendix/attn_map_2.pdf}
%     \caption{More examples of cross-attention maps.}
%     \label{fig:attn_map_2}
% \end{figure}

\clearpage
\subsection{Post-Generation Editing: Holistic View}

\begin{figure}[ht]
    \centering
    \includegraphics[width=1.0\columnwidth]{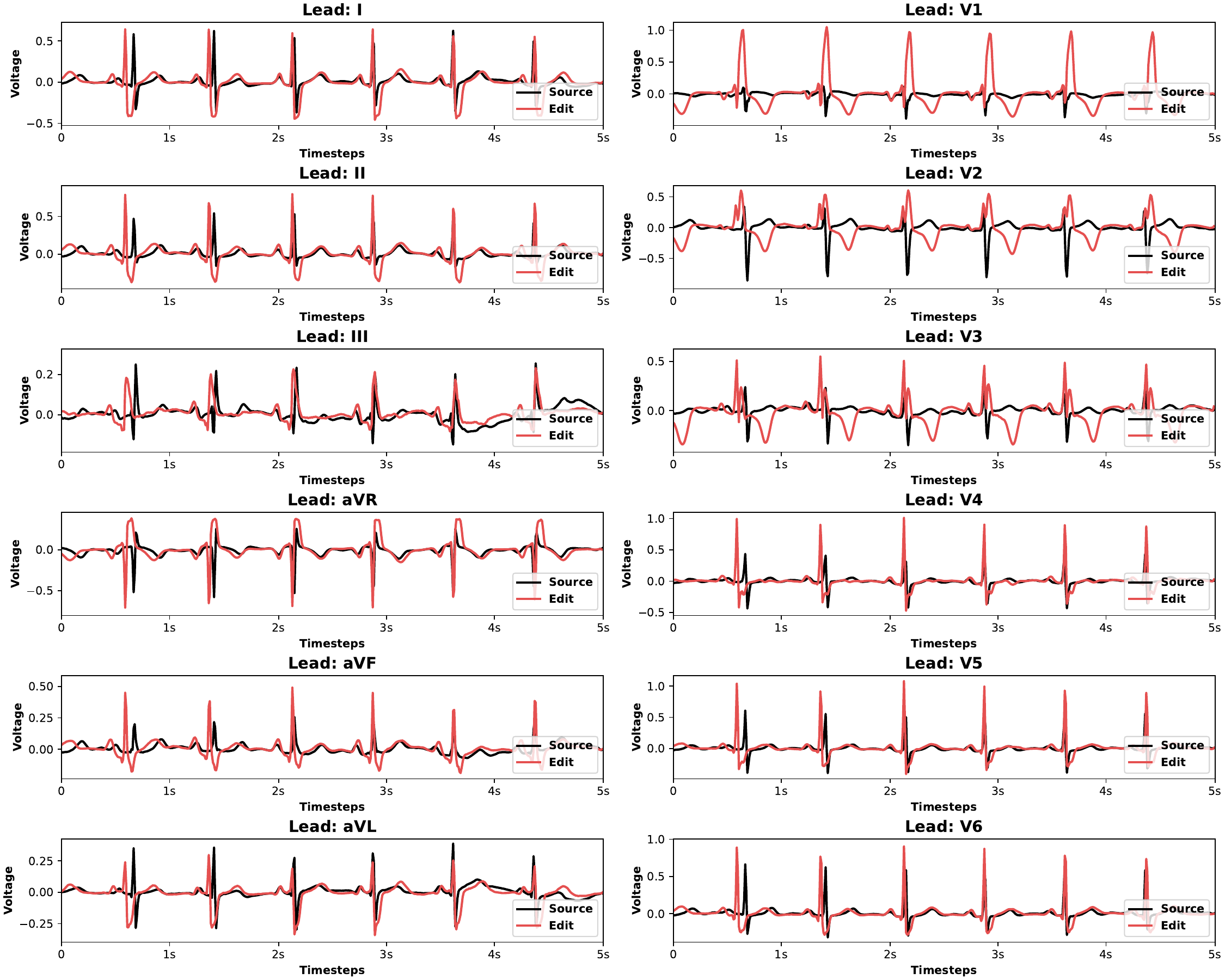}
    \caption{\textbf{Post-generation editing example.}}
    \label{fig:ecg_edit_all}
\end{figure}

In this section, we showcase the usage of prompt-to-prompt editing in \modelname\ by a typical cardiac disease \textit{Right Bundle Branch Block (RBBB)}.
The source ECG digital twin is generated under clinical tokens of \textit{sinus rhythm, Low QRS voltages in precordial leads, abnormal ecg} in target cardiac condition, and we add a new report token of \textit{Right bundle branch block} to it by performing post-generation ECG editing described aforementioned.
As illustrated in Fig.~\ref{fig:ecg_edit_all}, the resulting signal exhibits clear RBBB patterns while maintaining consistency with the original signal in terms of phase and morphology for regions unrelated to the new condition. Notably, all 12 leads exhibit coherent and semantically meaningful alterations, demonstrating the holistic nature of the modification. This capability for post-generation editing arises naturally from the Cardiac Condition Pathway and offers an additional layer of control over the generation process. It opens up new opportunities for applications such as causal analysis or being an interactive tools for cardiology education, both of which are exciting directions for future exploration.

\clearpage
\newpage
\section{Personalized ECG Generation Case Studies}
\label{app:case}

\begin{figure}[h!]
    \centering
    \includegraphics[width=0.85\columnwidth]{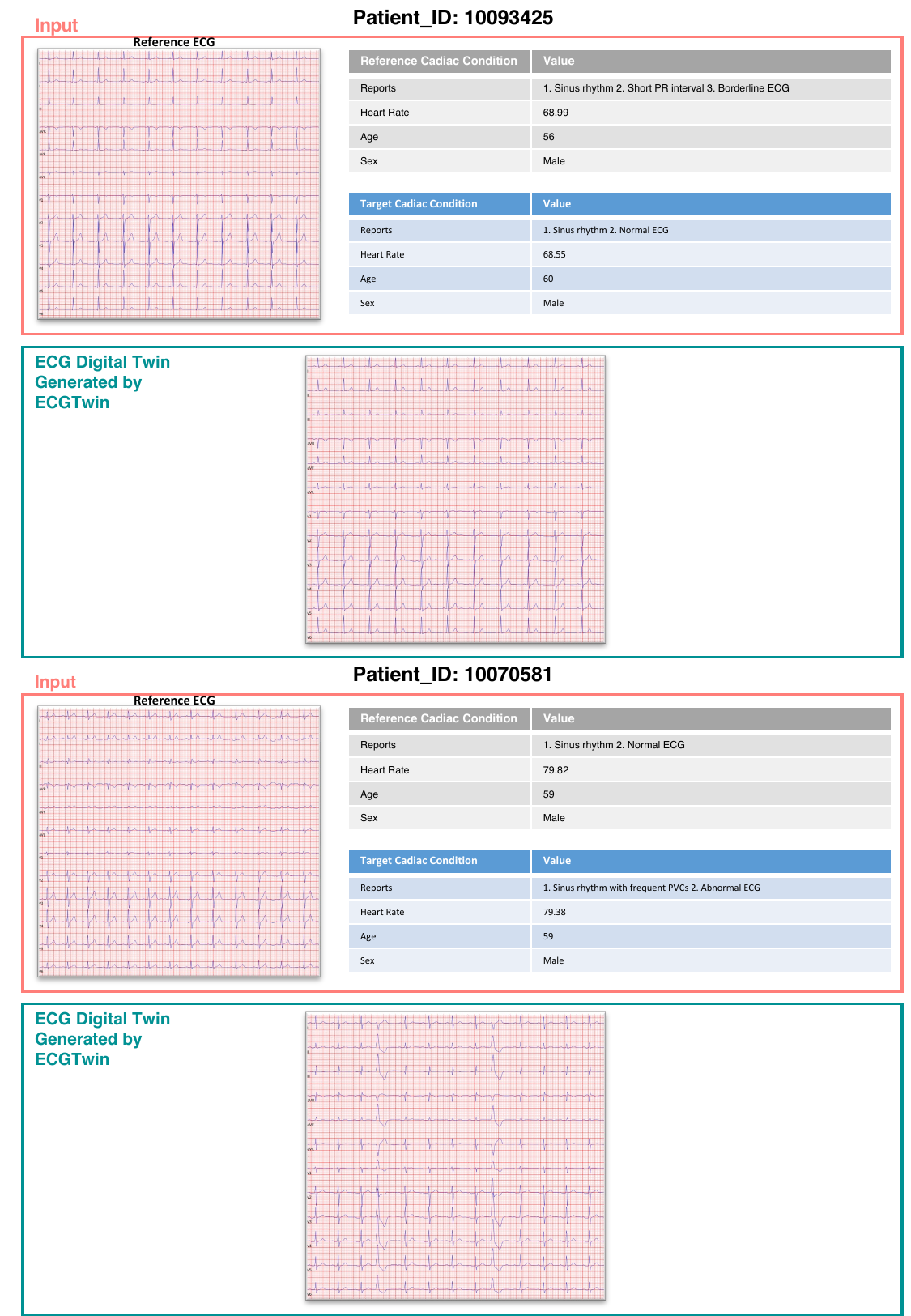}
    % \caption{Case Studies.}
\end{figure}

\begin{figure}[ht]
    \centering
    \includegraphics[width=0.85\columnwidth]{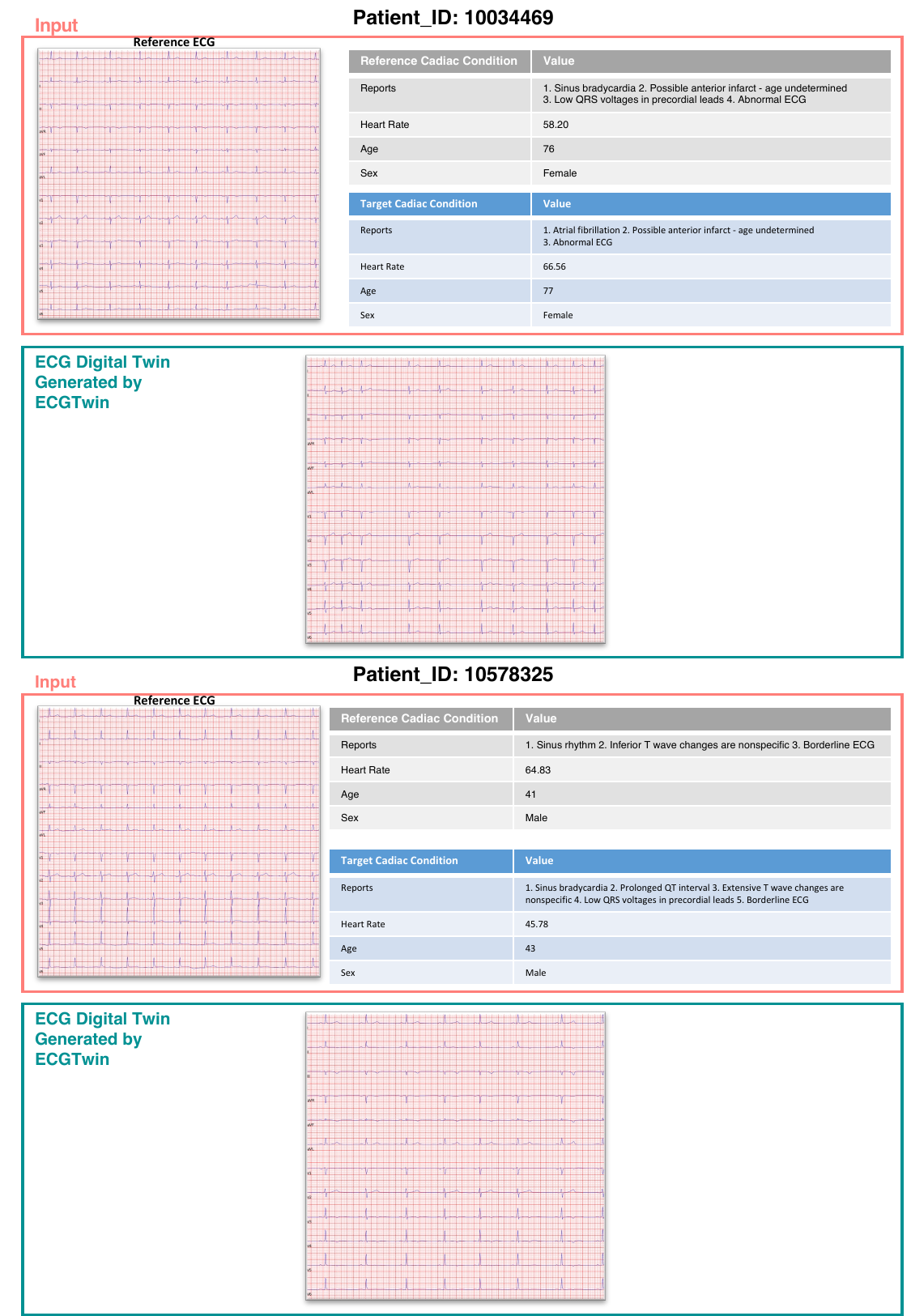}
    % \caption{Case Studies.}
\end{figure}

\begin{figure}[ht]
    \centering
    \includegraphics[width=0.85\columnwidth]{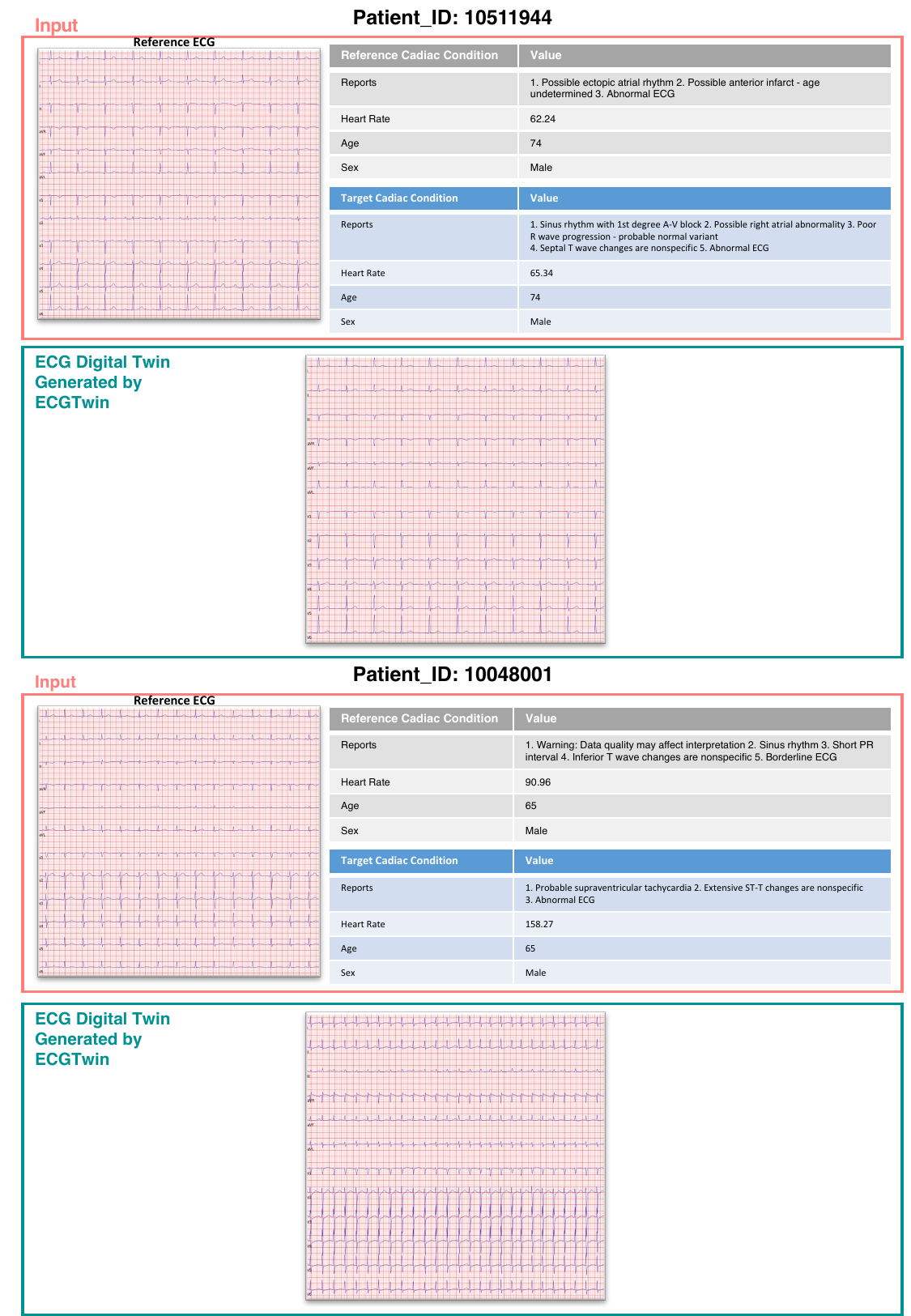}
    % \caption{Case Studies.}
\end{figure}

%%%%%%%%%%%%%%%%%%%%%%%%%%%%%%%%%%%%%%%%%%%%%%%%%%%%%%%%%%%%%%%%%%%%%%%%%%%%%%%
%%%%%%%%%%%%%%%%%%%%%%%%%%%%%%%%%%%%%%%%%%%%%%%%%%%%%%%%%%%%%%%%%%%%%%%%%%%%%%%

\end{document}